\documentclass[letterpaper, 10 pt, conference]{ieeeconf}  

\IEEEoverridecommandlockouts                              

\let\labelindent\relax

\usepackage{graphics, xcolor} 
\usepackage{tabularx}
\usepackage{amsmath, stmaryrd} 
\usepackage{cite}
\usepackage{diagbox}
\usepackage{multirow}
\usepackage{amsmath,amssymb,amsfonts}
\usepackage{newtxtext,newtxmath}
\usepackage{hyperref}
\usepackage{algorithmic}
\usepackage[ruled,linesnumbered,vlined]{algorithm2e}
\usepackage{graphicx}
\usepackage{textcomp}
\usepackage{xcolor}
\usepackage{balance}
\usepackage{enumitem} 
\usepackage{soul}
\usepackage{verbatim}
\makeatletter
\newcommand{\reducedplus}{\mathpalette\reduced@plus\relax}
\newcommand{\reduced@plus}[2]{%
  \sbox6{$\m@th#1+$}%
  \sbox8{\scalebox{0.875}{\copy6}}%
  \dimen@=\dimexpr(\wd6-\wd8)/3\relax
  \raisebox{\dimen@}{\box8}%
}
\newcommand{\boxoperation}[2][\mathbin]{%
  #1{\mathpalette\box@operation{#2}}%
}
\newcommand{\box@operation}[2]{%
  \ooalign{$\m@th#1\boxempty$\cr\hidewidth$\m@th#1#2$\hidewidth\cr}%
}

\def\secref#1{Sec.~\ref{#1}}
\def\figref#1{Fig.~\ref{#1}}
\def\tabref#1{Tab.~\ref{#1}}
\def\eqref#1{Eq.~(\ref{#1})}

\usepackage{fancyhdr}
\fancypagestyle{withfooter}{
  
  \fancyhead[L]{}
  \fancyhead[R]{}
  \fancyfoot[C]{\footnotesize Accepted to the IEEE ICRA Workshop on Field Robotics 2024}
}

\SetKwIF{Upon}{}{}{upon}{then}{}{}{}

\title{\LARGE \bf
A Sonar-based AUV Positioning System for Underwater Environments with Low Infrastructure Density
}

\author{Emilio Olivastri, Daniel Fusaro, Wanmeng Li, Simone Mosco, and Alberto Pretto
\thanks{This work was supported by the University of Padova under Grant UNI-IMPRESA-2020-SubEye.}
\thanks{The authors are with the Department of Information Engineering, University of Padova, Italy. Email: { \{olivastrie, fusarodani, wanmeng.li, moscosimon, alberto.pretto\}@dei.unipd.it}}
}
\begin{document}

\maketitle
\thispagestyle{withfooter}
\pagestyle{withfooter}

\begin{abstract}
The increasing demand for underwater vehicles highlights the necessity for robust localization solutions in inspection missions. 
In this work, we present a novel real-time sonar-based underwater global positioning algorithm for AUVs (Autonomous Underwater Vehicles) designed for environments with a sparse distribution of human-made assets. 
Our approach exploits two synergistic data interpretation frontends applied to the same stream of sonar data acquired by a multibeam Forward-Looking Sonar (FSD).
These observations are fused within a Particle Filter (PF) either to weigh more particles that belong to high-likelihood regions or to solve symmetric ambiguities.
Preliminary experiments carried out on a simulated environment resembling a real underwater plant provided promising results.
This work represents a starting point towards future developments of the method and consequent exhaustive evaluations also in real-world scenarios.

\end{abstract}

\section{Introduction}

\begin{figure}[t!]
   \centering
      \centering
   \begin{minipage}[b]{0.8\linewidth}
      \includegraphics[width=\linewidth]{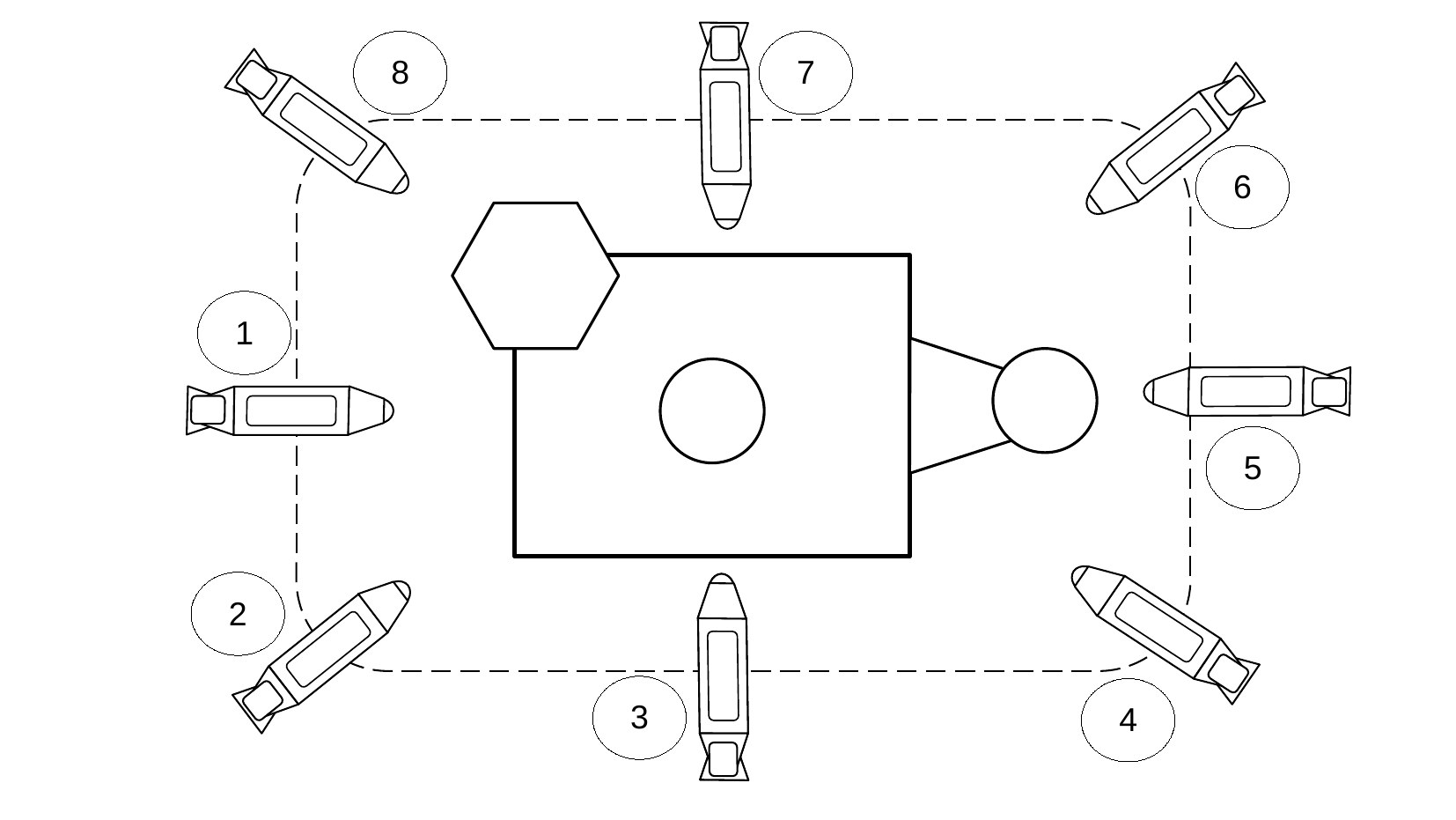}
   \end{minipage}\hfill
   \begin{minipage}[h]{0.8\linewidth}
      \includegraphics[width=\linewidth]{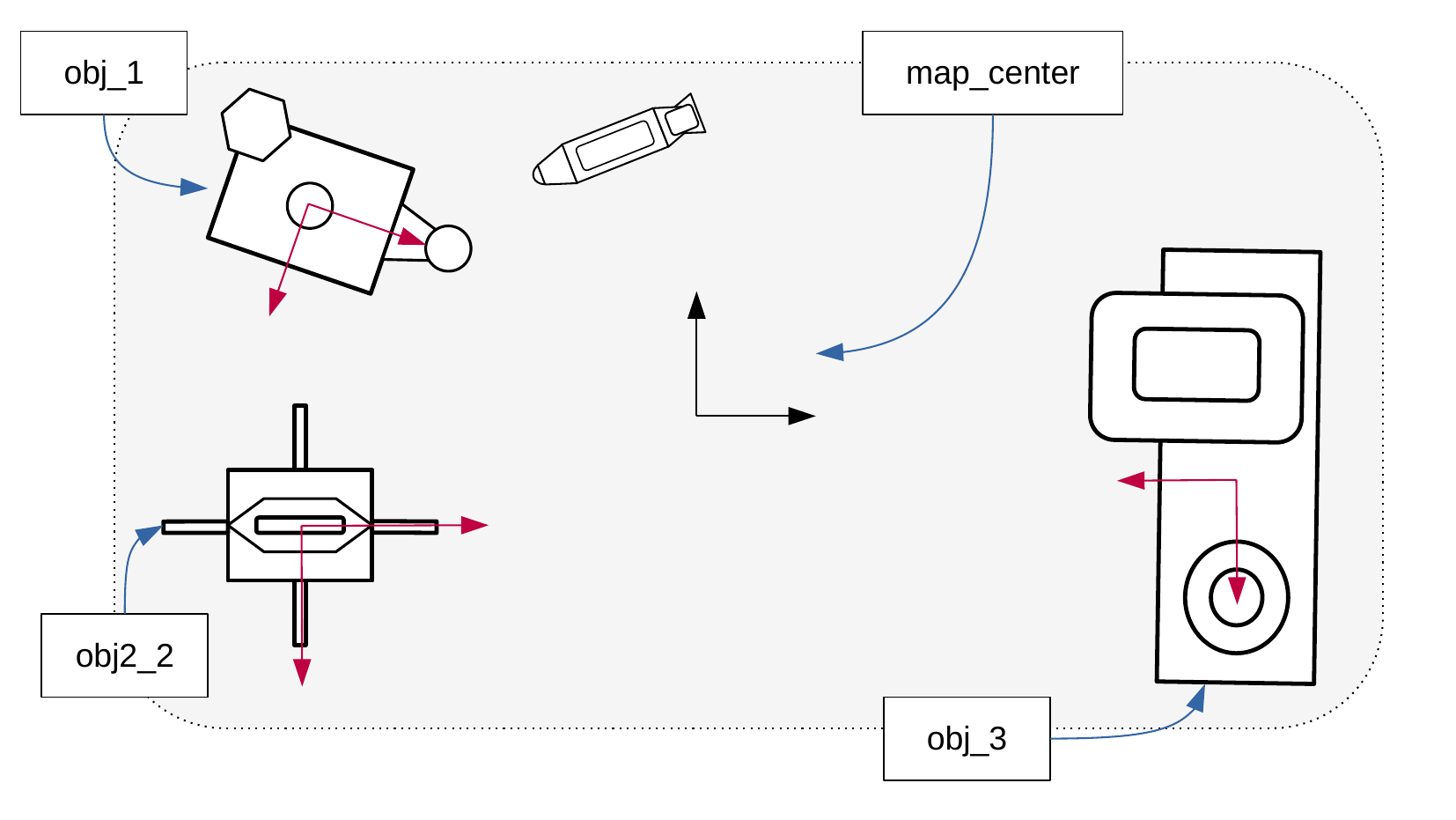}
   \end{minipage}\hfill
   \caption{The top image shows a typical trajectory (poses 1 to 8) followed by an AUV while inspecting a human-made asset, which was imitated during the virtual experiments. The bottom image shows a graphical representation of a map of an underwater plant. Approximate 3D CAD models of each "object" (i.e., asset) included in the map are also provided.}
   \label{fig:sonar_desc}
   \vspace{-5mm}
\end{figure}

In recent decades, there has been a rapid growth in the number of underwater plants, consequently increasing the demand for underwater vehicles capable of performing inspection and maintenance operations (e.g., \figref{fig:sonar_desc}, top). To this end, industries are trying to shift from Remotely Operated Vehicles (ROVs) to Autonomous Underwater Vehicles (AUVs). The Inertial Navigation System (INS) that equips such vehicles usually serves as the primary sensor for underwater navigation. However, INS-based dead reckoning navigation inevitably tends to accumulate drift over time, significantly limiting application scopes, even with high-end systems. On the other hand, vision-based autonomous navigation in underwater environments presents challenges due to factors such as light absorption and water turbidity, limiting the effectiveness of optical sensors. In fact, the most commonly used exteroceptive sensor in underwater operations is the sonar which is commonly not affected by these limitations.

The specificity of sonar data, however, introduces important new challenges that require ad-hoc methods and models.
As a result, traditional computer vision algorithms, which perform well on optical images, prove to be unreliable when applied to acoustic images. Instead, convolutional neural networks (CNNs) are capable of learning effective features from sonar images, but they require a large amount of data, which is often difficult to obtain due to the lack of publicly available datasets.
To address this constraint, synthetic data generation could prove to be an affordable and effective solution.
On the other hand, bridging the gap between synthetic and real scenarios remains an open challenge that still prevents the direct exploitation of synthetic data in real-world underwater applications.

In this paper, we introduce a novel underwater global positioning system for AUVs suitable for large-scale, sparsely structured environments that include few artificial infrastructures.
Our method integrates incoming data from both INS and a multibeam FSD (Forward-Looking Sonar), and estimates and updates the global pose of the AUV with respect to a provided map of a possibly large-scale environment. It requires as input an approximate map of the seabed (e.g., \figref{fig:sonar_desc}, bottom) and approximate 3D CAD models of each asset included in the map.  
To ensure reliable positioning, we integrate two complementary observation models: an underwater object detector and a place recognition module, both based on recent versions of the YOLO object detector \cite{jocher2022ultralytics}.
The fusion of the different observation models helps eliminate the ambiguity inherent in sonar measurements. 
We report preliminary experiments performed inside a simulated environment built with the Gazebo\footnote{https://gazebosim.org} simulator and the DAVE Aquatic Virtual Environment (DAVE) \cite{zhang2022dave} plugin for multibeam sonar simulation. This work represents the first step towards an effective sonar-based AUV positioning system that will be expanded and exhaustively evaluated in both simulated and real environments. Bridging the gap between synthetic model train and real-world deployment represents another challenge that we aim to address in future developments.\\

Our contributions can be summarized as follows:
\begin{enumerate}[label=(\roman*)]
    \item An object-oriented two-step place-recognition front-end, that builds upon a popular object detector to recognize both the relative position and the orientation of the sensor with respect to specific underwater assets; 
    \item A Particle Filter-based positioning back-end, that integrates the two heterogeneous observations provided by the front-end to localize the AUV inside the area of interest;
    \item A preliminary evaluation of the proposed method. The evaluation is performed in virtual settings that resemble closely real underwater operations.
    \normalsize
\end{enumerate}

\section{Related Work}
\label{sec:rel_works}

\subsection{Underwater AUV Localization}
Typical underwater localization methods exploit measurements from stationary acoustic transponder beacons distributed in the seabed \cite{4724364,YUAN2022103037} or Ultra-short baseline acoustic positioning (USBL) systems installed on both the AUV and a support vessel or nearby buoys \cite{5278162}. The landmark-based map of the beacons or buoys is known in advance or estimated as the solution to a SLAM (Simultaneous Localization And Mapping) problem (e.g., \cite{4089076}). Other approaches use natural landmarks (i.e., salient features) extracted from sonar or optical images \cite{li2018pose,rahman2022svin2}. Another class of methods exploits measurements of seabed topography to localize the AUV with respect to a given bathymetric map \cite{soton428141,TENG2020215}. Similarly, the presence of human-made infrastructures, knowing their locations, can be used to locate the AUV \cite{5278196,Kim2014,mcconnell2022overhead}. In \cite{5278196} authors proposed a sonar-based localization approach that uses a Particle Filter to match sonar data with a volumetric (possibly partially unknown) map of the underwater environment. A vision-based localization method for structured underwater environments is presented in \cite{Kim2014}, where authors proposed to exploit a template-based object detection technique and apply it to the Monte Carlo localization (MCL) algorithm. In \cite{mcconnell2022overhead} the authors proposed a localization algorithm for AUVs moving within a marina or a ship's pier. Localization is obtained by matching aerial images of the infrastructures with overhead images synthetically generated from sonar images by using a CNN. 
Our method falls into this last class of methods and, being object-detection-based, recalls the technique proposed in \cite{Kim2014} but, compared to this, we rely on sonar data and exploit a deep learning-based object detector and an orientation classification.
Most of the approaches presented above rely on probabilistic techniques such as Particle Filters, nonlinear versions of the Kalman filter such as EKF \cite{books/daglib/0014221}, or pose-graphs \cite{5681215}. A recent review of AUV localization techniques can be found in \cite{10.1007/s41315-021-00215-x}.
\subsection{Object Detection in Sonar Images }
Our approach is based on object detection performed directly on sonar images. Early attempts for sonar image object detection were based on handcrafted features. For example, in \cite{5475293} a template matching algorithm that takes into account both the echoes and the shadows produced by the object for distinguishing foreground from background objects is presented. In \cite{6608106} authors consider local matches over time to take into account also variations in the responses due to viewpoint changes.
With recent improvements in sonar technologies, convolutional neural networks (CNNs) are starting to achieve good results in sonar object detection tasks. 
To this end, state-of-the-art object detector and segmentation models used for optical images were adapted to sonar images, among others Faster RCNN\cite{karimanzira2020object}, Mask RCNN\cite{fan2021detection}, and YOLO\cite{7778702}. Recently \cite{rs13183555} included attention layers \cite{vaswani2017attention} in a YOLO-based architecture to enhance detection capability while in \cite{neves2020rotated} authors proposed an object detector composed of two CNNs that output both the object's position and its rotation.

\subsection{Sonar-based Place Recognition}
Place Recognition is the challenging task of recognizing the location of a given query image, assuming we have (or are able to build) an image database of our workspace. One possible solution to solve this task is through the use of descriptors, that can be classified into two categories: local and global. The former relies on techniques such as SIFT \cite{790410}, SURF \cite{10.1007/11744023_32}, and VLAD \cite{5540039} to extract local feature descriptors. Typically, for place recognition with descriptors, it's preferred to use the latter technique, because of its low overhead, which focuses on the global scene by predefining a set of key points within the image and subsequently converting the local feature descriptors of these key points into a global descriptor during the post-processing stage. For instance, the widely utilized Gist \cite{OLIVA200623} and HoG \cite{1467360} descriptors are frequently employed for place recognition across various contexts.\\
Still, the main problem is that the handcrafted features that used to work for optical images are not as effective for sonar images due to the challenging environmental conditions. \cite{7759205} first proposed to use CNN's learned features as descriptors. Expanding upon this, \cite{8260693} introduces a siamese CNN architecture to predict similarity scores among underwater sonar images. In \cite{8614109} presents a variant of PoseNet relying on the triplet loss commonly used in face recognition tasks. Particularly, they utilize an open-source simulator \cite{CERQUEIRA2020} to synthesize forward-looking sonar (FLS) images, from which the network learns features and demonstrates strong performance on real-world sonar datasets. Inspired by these advancements, \cite{9267885} utilizes the triplet loss combined with ResNet to learn the latent spatial metric of side-scan sonar images. Additionally, \cite{8793550} employs a fusion voting strategy using convolutional autoencoders to facilitate unsupervised learning of salient underwater landmarks. Recently, we proposed \cite{donadi2023improving} a sonar image descriptor based on ResNet and a random Gaussian projection layer that builds upon a customized synthetic data generation procedure.

\section{Methodology}
\label{sec:methodology}
\subsection{Autonomous Inspection of Underwater Structures}
\label{sec:assumptions}
The typical underwater mission we address consists of two main operations: reaching and inspecting the asset. Inspection is carried out by performing an almost rectangular trajectory while always facing the object, as depicted in \figref{fig:sonar_desc}. 
Once one inspection is finished, the next action can be to move to the next asset to be inspected or return to the base station.
The most common sensor configuration for this type of operation is the following:
\begin{itemize}
    \item Forward-looking sonar;
    \item INS/DVL positioning system;
    \item Camera;
    \item Altimeter.
\end{itemize}
Given the described working conditions, in this work, the following assumptions and simplifications will be made:
i) the vehicle speed during the inspection is low, enabling the adoption of a simple, approximated motion model such as the constant velocity model; ii) the AUV is kept almost perfectly horizontal during the missions, allowing us to approximate and fix the values of both roll and pitch to zero; iii) the altimeter gives reliable information relative to the global altitude of the vehicle. Fusing the altimeter information with the motion estimation enables a decoupled robust altitude estimation. These approximations enable us to focus primarily on estimating three degrees of freedom (DoFs), respectively the robot's 2D location $[x, y]$ and its yaw $\theta$, significantly reducing the complexity of the localization algorithm.\\
Our approach requires a 2D map (planimetry) $\mathcal{M}$  of the underwater plant reporting the positions of the installed human-made structures (assets), plus the related (simplified) CAD models of each asset. The map $\mathcal{M} = \{ ^{j}\mathbf{L} \in \text{SE}(3) | j = 1,...,K \} $ is defined as a set of $K$ poses $^{j}\mathbf{L}$ that represents the pose of the asset $j$ with respect to the center of the map, an example can be seen in \figref{fig:sonar_desc}.

\subsection{Sonar Preliminaries}

\begin{figure}[t!]
    \centering
    \includegraphics[width=0.4\textwidth]{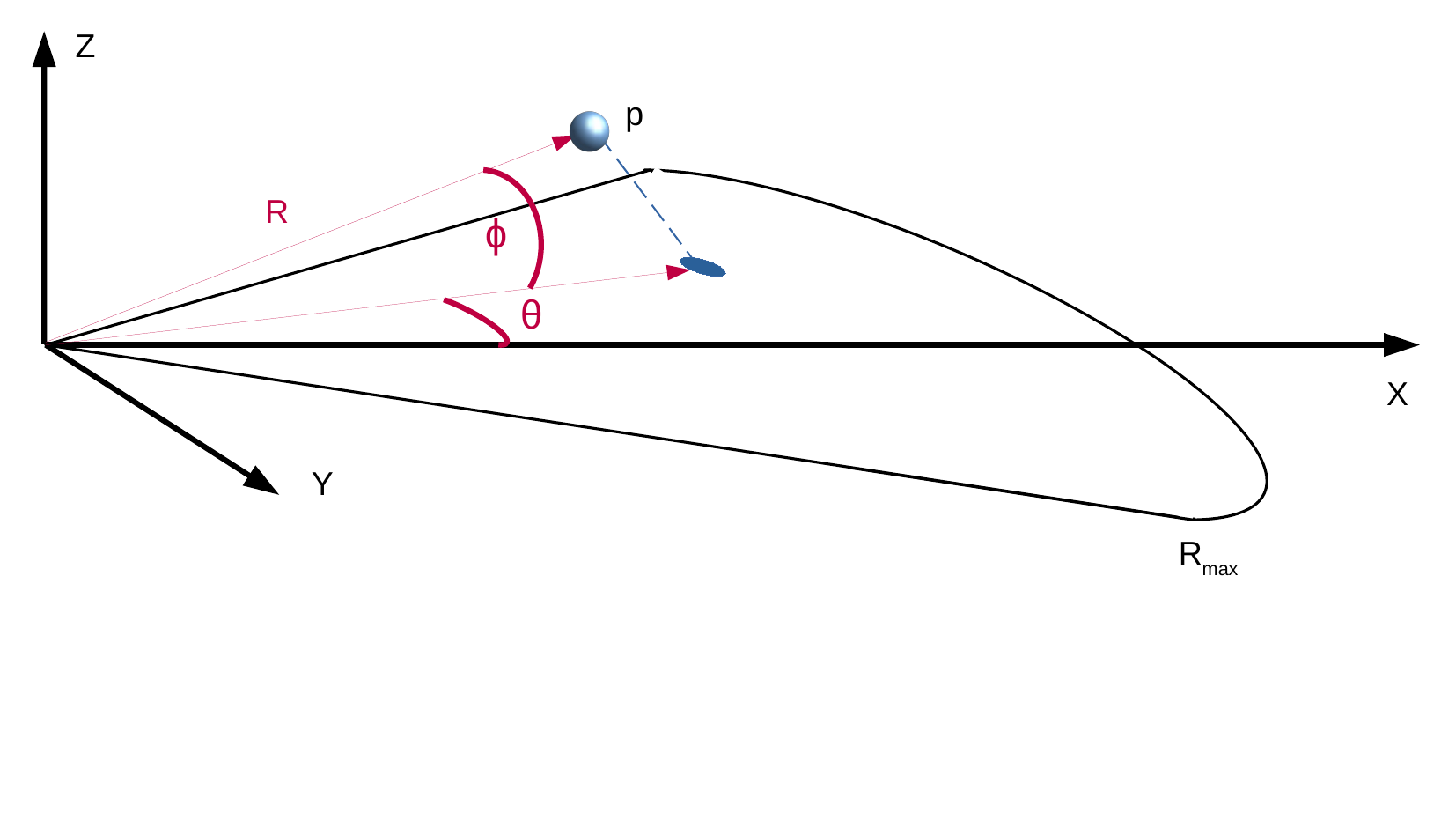}
    \vspace{-10mm}
    \caption{Example of the sonar's projection model. Here it is evident that all the points that live in the arc $\phi$ will be projected in the same point.}
    \label{fig:sonar_projection}
    \vspace{-5mm}
\end{figure}
It is useful to introduce some notions regarding the used sensor i.e., a Forward-Looking Sonar (FLS). In particular, we are interested in how a point $\mathbf{p} = [x_s, y_s, z_s]$ in the sonar's cartesian reference frame can be projected in the sonar image\cite{li2018pose} with coordinates $\mathbf{c}=[u, v]$. Given the relation between cartesian and spherical coordinates in \eqref{eq:polar_2_cartesian} (see also \figref{fig:sonar_projection}):
\begin{equation}
\label{eq:polar_2_cartesian}
\begin{bmatrix}
x_s\\
y_s\\
z_s
\end{bmatrix}  = \begin{bmatrix}
                  R cos\theta cos\phi\\
                  R sin\theta cos\phi\\
                  R sin\phi
                  \end{bmatrix}    
\end{equation}
we can reverse it to switch to spherical coordinates \eqref{eq:cartesian_2_polar}, which is a more convenient representation when performing the sonar projection: 

\begin{equation}
\label{eq:cartesian_2_polar}
\begin{bmatrix}
R\\
\theta\\
\phi
\end{bmatrix}  = \begin{bmatrix}
                  \sqrt{x_s^2 + y_s^2 + z_s^2}\\
                  tan^{-1}\frac{x_s}{y_s}\\
                  tan^{-1}\frac{z_s}{ \sqrt{x_s^2 + y_s^2}}
                  \end{bmatrix}    
\end{equation}

To finally project the point $\mathbf{p}$ to pixel $\mathbf{c}$ in a sonar image of size $[W, H]$, that are respectively its width and height, it is needed to apply the following formula:
\begin{equation}
\label{eq:sonar_projection}
\begin{bmatrix}
u\\
v
\end{bmatrix}  = \begin{bmatrix}
                  \beta \frac{y_s}{cos\phi} + \frac{W}{2}\\
                  \beta (R_{max} - \frac{x_s}{cos\phi})
                  \end{bmatrix}    
\end{equation}
where $\beta$ is the scaling factor that puts in relation the physical and image space, by taking into account the parameters of the FLS, such as its range max $R_{max}$ and its field of view $FoV$ (expressed in radians), and the dimensions of the sonar image in the following manner:
\begin{equation}
\label{eq:factor_conversion}
\beta = \frac{W}{2R_{max} sin{\frac{FoV}{2}}}.
\end{equation}
Another important note, relative to the ambiguity of $\phi$, needs to be highlighted when using this projection model. When projecting to the sonar image we lose the height information, which means that all the points that live in the arc defined by $[R, \theta]$ ( like it is shown in \figref{fig:sonar_projection} ), will be projected into the same pixel.

\subsection{Underwater Particle Filter}
The localization pipeline is based on the particle filter method\cite{djuric2003particle}, which models the posterior distribution of the robot's pose $\mathbf{X}_t \in \text{SE}(2)$ at the time $t$ as a set of random samples, called particles, drawn from such distribution. The set of particles is defined in the following way:
\begin{equation}
    \label{eq:pf_set}
    \mathcal{S}_t = \{ \mathbf{X}_{t}^{i} \in \text{SE}(2) | i = 1,...,N \},
\end{equation}
where each particle $\mathbf{X}_{t}^{i}$ represents a robot's pose hypothesis, while $N$ denotes the number of total particles. 
The particle filter updates the set of samples $\mathcal{S}_{t}$ through the following three steps:
\begin{itemize}
    \item Sampling: $\mathcal{S}_{t}$ is generated from the previous $\mathcal{S}_{t-1}$ by sampling from the proposal distribution $p(\mathbf{X}_{t}^{i} | \mathbf{X}_{t-1}^{i}, \mathbf{u}_t)$, which is the motion model of the AUV, where $\mathbf{u}_t$ is the control input received at time $t$.
    \item Re-weighting: A weight $w^i_{t}$ is assigned to each particle in $\mathcal{S}_t$, and it accounts for the error of using the proposal distribution, instead of the target distribution(real posterior distribution), to generate particles.
    In robot localization the weight $w^i_{t} \propto p(\mathbf{z}_{t} | \mathbf{X}_{t}^{i}, \mathcal{M}) $ is proportional to the observation model, where $\mathbf{z}_{t}$ is the measurement received at the time $t$. 
    \item Resampling: Final step that generates a new set of particles $\mathcal{S}_{t}{'}$, where each particle is sampled with probability $w^i_{t}$. This step leads to maintaining the most likely particles while eliminating the least likely ones. 
\end{itemize}

\subsection{Sonar Asset Detector}
\label{sec:object_detector}
\begin{figure}[t!]
    \centering
    \includegraphics[width=0.4\textwidth]{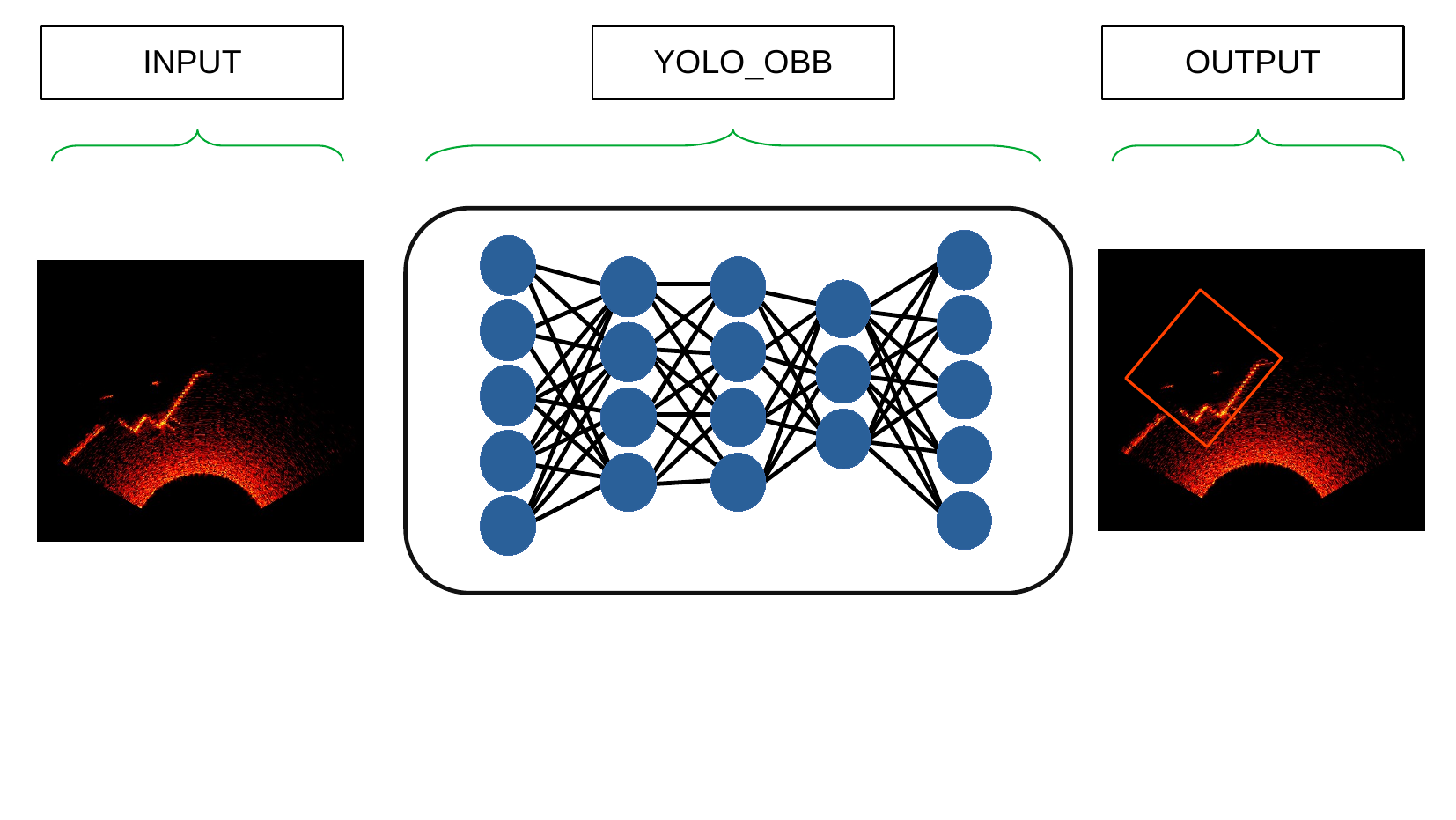}
    \vspace{-10mm}
    \caption{Example of the YoloOBB detection pipeline using a sonar image produced utilizing the simulator Dave \cite{zhang2022dave}. The OBBs produce a more informative measurement with respect to the classic BBs.} 
    \label{fig:example_detection}
    \vspace{-5mm}
\end{figure}
\begin{figure}[t!]
   \centering
   \begin{minipage}[t]{\linewidth}
      \centering
      \includegraphics[width=0.5\textwidth]{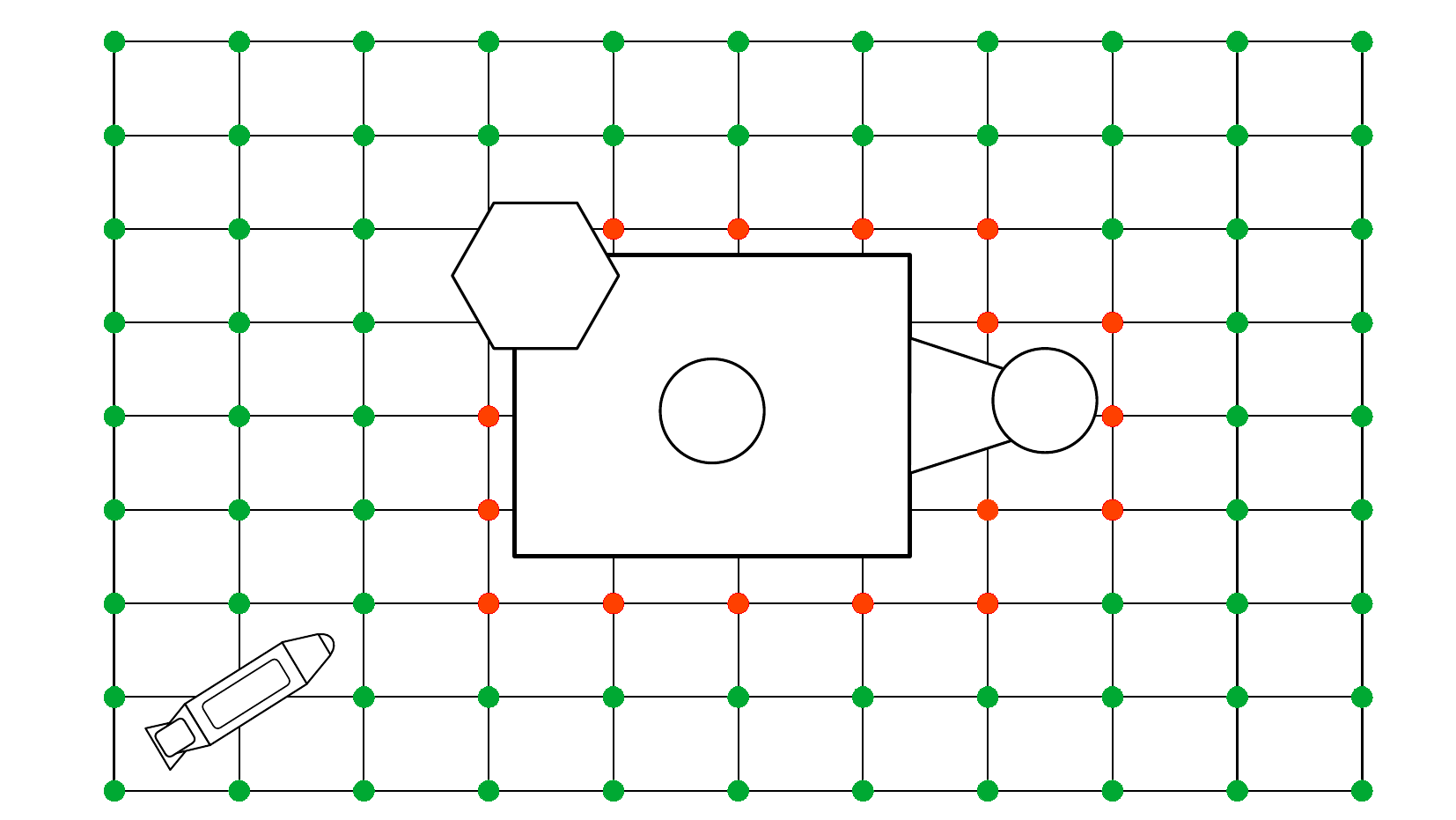}
   \end{minipage}\hfill
   \caption{Representation of the 2D grid centered on the asset used to produce the dataset for training the two different networks. The green dots represent valid poses in which there is no collision, while the red dots represent poses to be avoided due to collisions.}
   \label{fig:dataset}
   \vspace{-3mm}
\end{figure}

In the following section, it will be described the first observation model used for assigning the weights $\mathbf{w}_t$ to the particles.
The sonar asset detector (SAD) module is composed of the Yolo-v5 \cite{jocher2022ultralytics} object detector plus an extension \cite{yang2020arbitrary} that is able to estimate also the orientation of the bounding boxes surrounding the object (Oriented Bounding Boxes, OBB for short). The goal of the module is to provide the 2D coordinates of the object $j$ in the sonar image, its class, and its orientation. The network was trained in a simulated environment that is representative of a real-world underwater plant. A sparse map, the CAD models of the objects were used for building the environment, while the plugin Dave\cite{zhang2022dave} simulated the water effects, the FLS, and its noise.\\
The dataset was generated using a similar strategy as the one used in \cite{donadi2023improving}: a 2D grid is generated around the object, as shown in \figref{fig:dataset}, and for each valid pose in the grid, three sonar images and corresponding OBBs were computed: one with the sonar facing the center of the object and the other two views were generated adding $\pm15^{\circ}$ to the central view. To compute the respective OBBs we need to first estimate, from the pose of the sonar $\mathbf{X}_s$ and the pose of the object $^{j}\mathbf{L}$:
\begin{equation}
    \label{eq:sonar_space}
    ^{j}\mathbf{L}_{s} = [\mathbf{X}_{s}]^{-1} \cdot {^{j}\mathbf{L}},
\end{equation}
that is the pose of the object $j$ in the reference frame of the sonar. The translation component of $^{j}\mathbf{L}_{s}$ corresponds to the point $\mathbf{p}$ used in \eqref{eq:polar_2_cartesian}, while from the rotation component we can extract the $\delta$, which represents the rotation of the bounding box. Finally, given the parameters of the sonar we are able to correctly estimate the center of the OBB using \eqref{eq:sonar_projection}.

The observation model that was formulated from this module is reported in Algorithm~\ref{alg:obs_yolo}. As soon as the sonar image $\mathbf{z}_t$ is received (line \ref{alg:obs_yolo}), it is given as input to the network (line \ref{alg:ass_detect}) in order to retrieve the center of the OBB $\mathbf{\Tilde{c}}_{j} = [\Tilde{u}_j, \Tilde{v}_j]$, it's rotation $\Tilde{\delta}$ and the class of the object detected. If the network returns a successful detection (line [\ref{alg:ass_next_step}-\ref{alg:ass_pose}]), then we can select the pose of the asset $j$ that will be projected in the sonar image of each particle (line [\ref{alg:obj_for}-\ref{alg:project_obj}]). That means that we obtain for each particle in the system the corresponding projection $\mathbf{\hat{c}}_{ij} = [\hat{u}_{ij}, \hat{v}_{ij}]$ and orientation $\hat{\delta}$ of the object $j$. Finally, the new weight for each particle $i$ (line \ref{alg:obj_reweight}) will be computed as follows:
\begin{equation}
    \label{eq:weight_obs1}
    w^i_t = w^i_t \cdot \texttt{exp}(-dist_{ij}),
\end{equation}
with the exponential factor $dist_{ij}$ (line \ref{alg:distance}) computed as:
\begin{equation}
    \label{eq:dist_obs1}
    dist_{ij} = \eta|| \mathbf{\hat{c}}_{ij} - \mathbf{\Tilde{c}}_{j}||^2 + ||\texttt{sin}(\hat{\delta}_{ij} - \Tilde{\delta}_j)||
\end{equation}
where $\eta$ is a normalization factor used for balancing the pixel error with the angle error. For the latter, we use the $||sin()||$ function because it naturally handles the fact that both $\Tilde{\delta}_j$ and $\Tilde{\delta}_j + \pi$ are valid solutions.

\begin{figure}[b!]
   \centering
   \begin{minipage}[b]{\linewidth}
      \centering
      \includegraphics[width=0.8\linewidth]{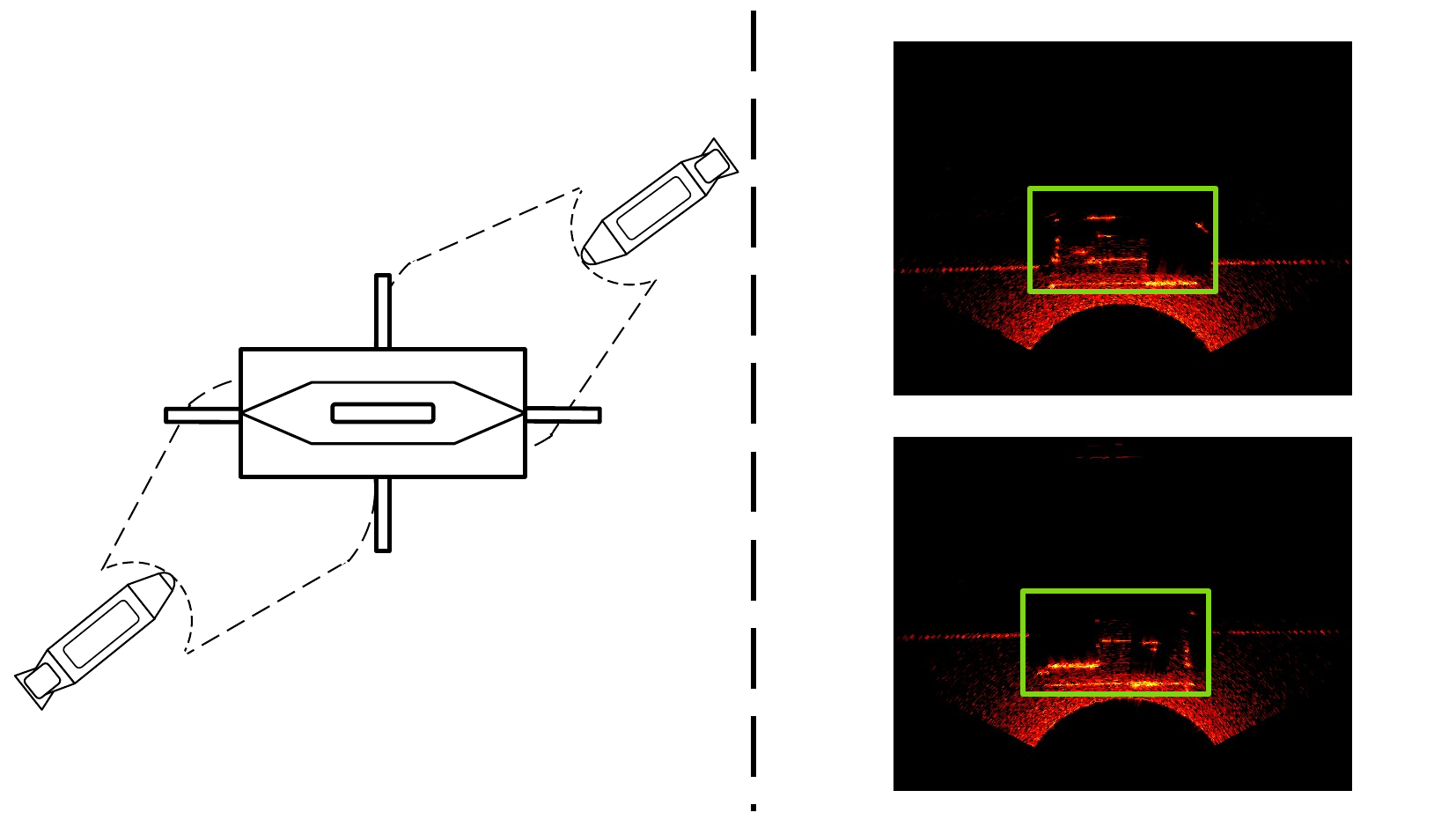}
   \end{minipage}\hfill
   \caption{On the left it is showcased a situation in which there is a symmetric underwater asset, and the two poses from which the sonar images computed will result to be almost the same. On the right, it is shown sonar images produced by the simulator in which the results of the detection are ambiguous.}
   \label{fig:ambiguity}
\end{figure}

When applying a projection following the camera model, the depth information of the image is lost. However, the projected object retains lots of features that make it possible to retrieve the 6D pose of the object. This doesn't happen when applying a projection following the sonar model, aside from losing the height information it also has poor resolution and lots of noise leading to the loss of important and discriminative features. Furthermore, the objects projected in the sonar images are typically symmetrical, making it so that the observation can be generated from two opposing views centered at the object as shown in \figref{fig:ambiguity}.
When computing the new weights of the object this fact is also taken into account, so after the update step both solutions are kept valid in the system.
\begin{algorithm}[h!]
\DontPrintSemicolon
\caption{Module Asset Detector Observation}\label{alg:obs_yolo}
\small
\Upon{receiving $\mathbf{z}_{t}$}
{
    $\mathbf{\Tilde{c}}_{j}, \Tilde{\delta}_j, j \leftarrow \texttt{YoloOBB}(\mathbf{z}_{t})$ \label{alg:ass_detect} \tcp{Detect object}
    \tcp{Successful detection}
    \If{ j is valid  } 
    {\label{alg:ass_next_step}
        $^{j}\mathbf{L} \leftarrow \texttt{getObjectPose}(j, \mathcal{M})$\; \label{alg:ass_pose}
        \For{$i\gets0$ \KwTo $N-1$ }
        {\label{alg:obj_for}
            $\mathbf{X}_{t}^i \leftarrow \texttt{getParticlePose}(j)$\;  \label{alg:particle_pose}
                $\mathbf{\hat{c}}_{ij}, \hat{\delta}_{ij} \leftarrow \texttt{projectOBB}(\mathbf{X}_{t}^i, {^{j}\mathbf{L}})$\; \label{alg:project_obj}
            $dist_{ij} \leftarrow \texttt{computeDistance}(\mathbf{\hat{c}}_{ij}, \hat{\delta}_{ij}, \mathbf{\Tilde{c}}_{j}, \Tilde{\delta}_j)$\; \label{alg:distance}
            $w_t^i \leftarrow w_t^i \exp(-dist_{ij})$ \label{alg:obj_reweight}
        }
    }
}
\end{algorithm}

\subsection{Place Recognition Module}
\begin{figure}[t!]
   \centering
   \begin{minipage}[b]{\linewidth}
   \centering
      \includegraphics[width=0.5\linewidth]{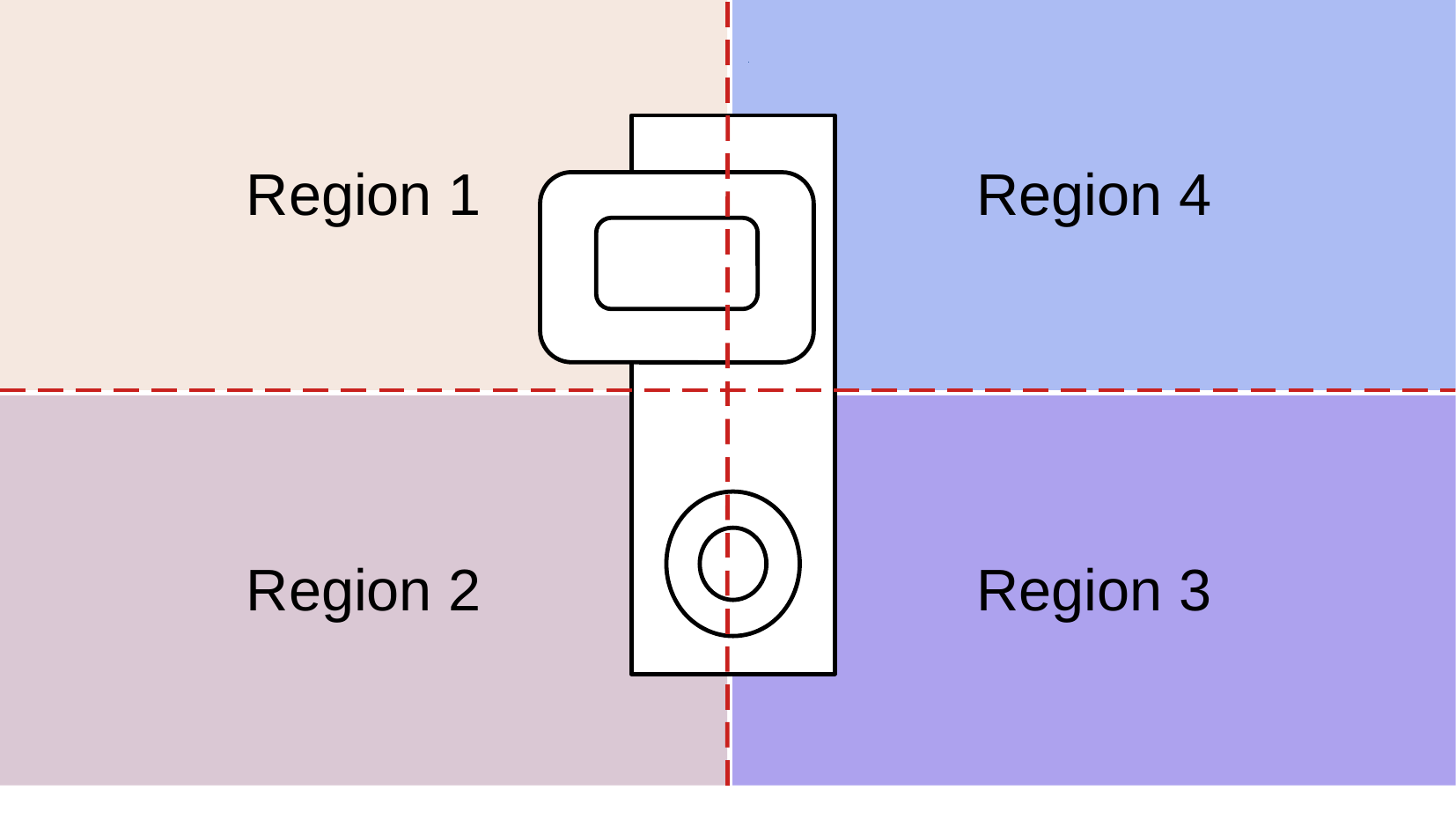}
   \end{minipage}\hfill
   \caption{Example of how the space around an object is divided in regions for the classification problem.}
   \label{fig:topological}
   \vspace{-5mm}
\end{figure}

When inspecting different objects in the underwater environment, the ambiguity of the SAD module will be eventually solved. But as there are scenarios in which the scene is populated with multiple assets, there are also occasions where an isolated object needs to be inspected.
In such cases, resolving the ambiguity could be not possible, especially if they are strongly symmetric, leading to an imprecision localization during the inspection.\\\
To solve this issue, a place recognition module (PRec) has been added to the system. We formulated the place recognition problem as a scene classification problem\cite{liu2018dictionary}, where it doesn't only classify which asset the robot is currently seeing, but also which of its sections it is inspecting. 
In this case, we only need to divide the assets into four regions, as shown in \figref{fig:topological}, because is it sufficient information to complement the SAD. 
In total, the number of classes that need to be learned are $4K + 1$, where the extra class is needed for the cases in which the robot is not observing anything.  
The same sonar images of the dataset used in \secref{sec:object_detector} are re-used here. The only difference is how the labels are produced. For each object $j$, other than the OBB annotation, when sampling the poses of the grid (\figref{fig:dataset}) it is also computed the region of belonging. To do so, first it is needed to take the pose $^{j}\mathbf{L}_{s}$ defined in \eqref{eq:sonar_space}, and invert it, in order to have the sonar in the reference frame of the object $j$. Then, we use the translation component of the new matrix to compute the $\theta^*$ using \eqref{eq:cartesian_2_polar}, which identifies the object's region in which the sonar is positioned at the moment.\\
The classifier used for the PRec module is Yolo-v8 (developed by the same authors of \cite{jocher2022ultralytics}), which was chosen because of its state-of-the-art performance and its ease in usability.\\
The observation model for this module has a similar strategy as the one defined for the SAD (see Algorithm~\ref{alg:obs_place}).
As soon as the sonar image $\mathbf{z}_t$ (line \ref{alg:obs_place}) is received, it is then fed to the network (line \ref{alg:ass_classify}) in order to retrieve both the object id $j$ and its relative section. If the robot is actually observing an asset (line [\ref{alg:next_step_2}]), then the weight $w_t^i$ of all the particles that fall in the detected region are increased by a factor $\alpha > 1.0 $ (line [\ref{alg:reweight_2}-\ref{alg:obs2_end}]).\\
This module alone would not be able to reach good localization accuracy on its own, since all the particles in said region have the same weight, so it would give as a pose estimate, on average, the center of the identified asset's region.

\begin{algorithm}[h!]
\DontPrintSemicolon
\caption{Place Recognition Module Observation}\label{alg:obs_place}
\small
\Upon{receiving $\mathbf{z}_{t}$}
{
    \tcp{Recognize region}
    $reg \leftarrow \texttt{PlaceRecognition}(\mathbf{z}_{t})$ \label{alg:ass_classify}\;
    \If{ reg is not None  } 
    {\label{alg:next_step_2}
        \For{$i\gets0$ \KwTo $N-1$ }
        {\label{alg:reweight_2}
            $\mathbf{X}_{t}^i \leftarrow \texttt{getParticlePose}(i)$\;
            \If{ $\texttt{isInRegion}(\mathbf{X}_{t}^i, reg)$  }
            {
                $w_t^i \leftarrow w_t^i \cdot \alpha$ \label{alg:obs2_end}
            }
        }
    }
}
\end{algorithm}

\begin{figure*}[t!]
    \centering
    \caption{The red and blu arrows, respectively, represent the ground truth pose and the estimated of the AUV in the 2D. The top left and top images showcase, respectively, the trajectories carried out by the vehicle during mission 1 and mission 2. The top left and top images showcase, respectively, a successful localization run and a failed one during mission 3.}
    \begin{minipage}[t]{0.40\textwidth}
        \centering
        \includegraphics[width=\textwidth]{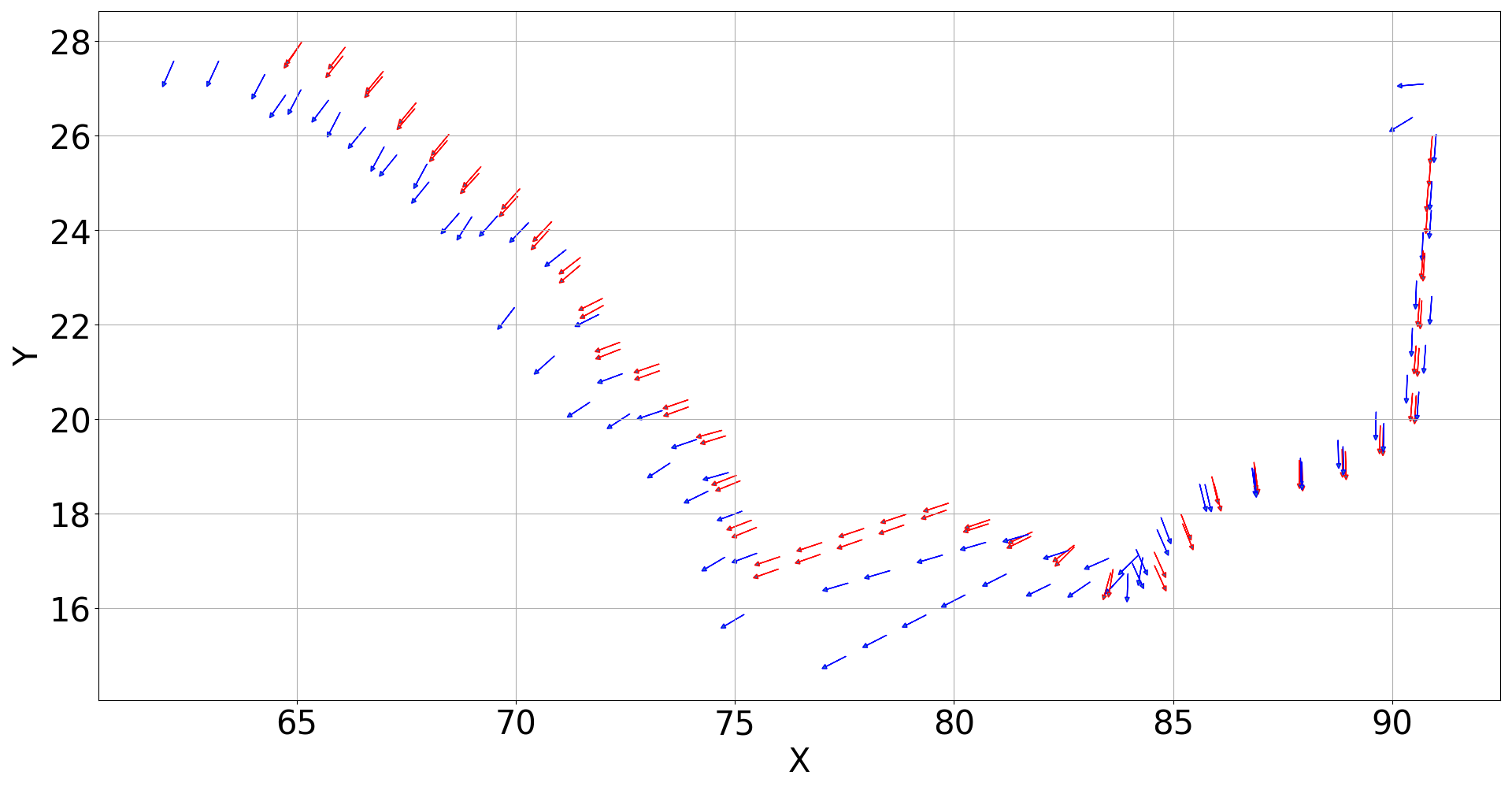}
    \end{minipage}
    \begin{minipage}[t]{0.40\textwidth}
        \centering
        \includegraphics[width=\textwidth]{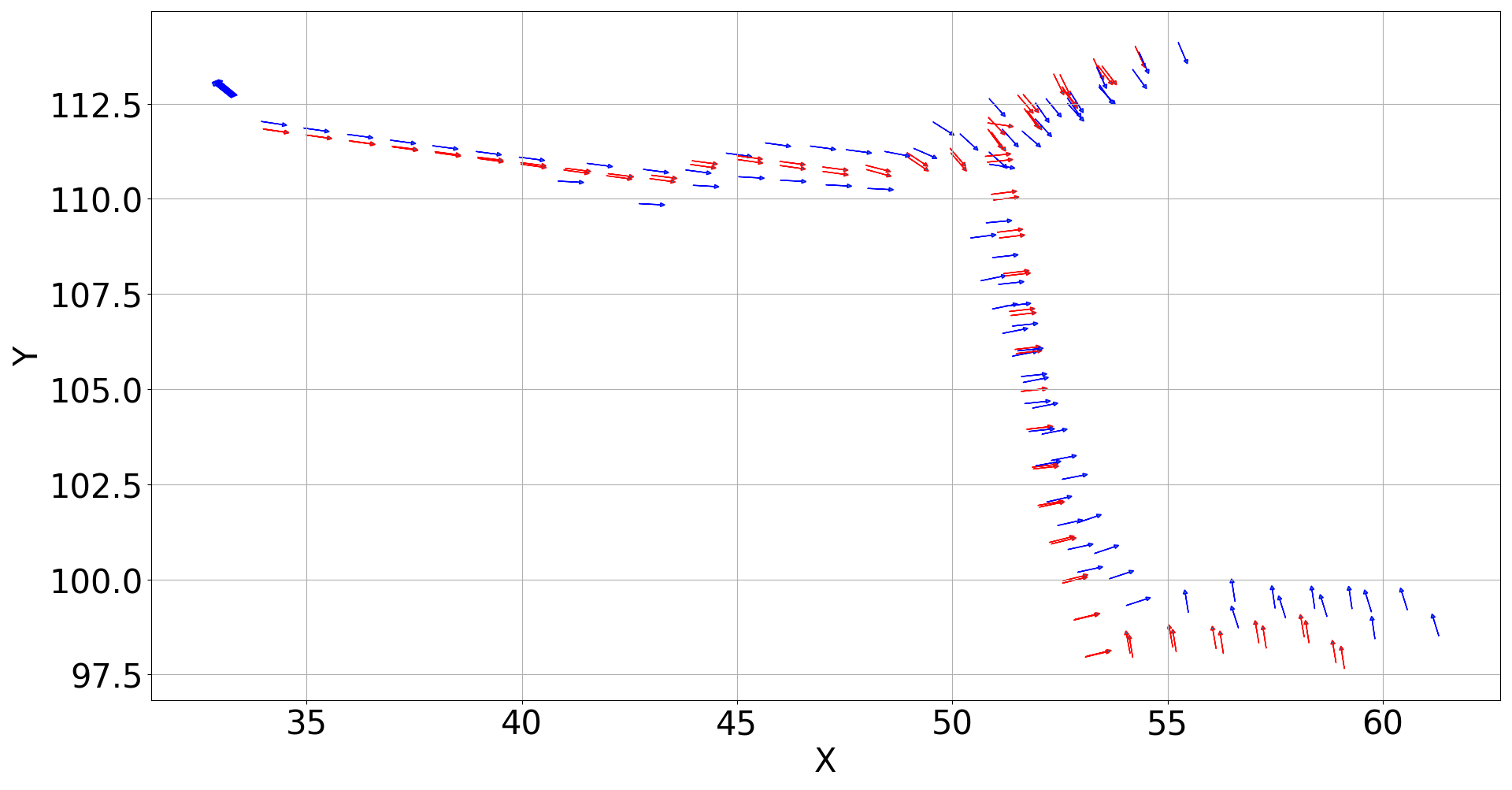}
    \end{minipage}
    \begin{minipage}[t]{0.40\textwidth}
        \centering
        \includegraphics[width=\textwidth]{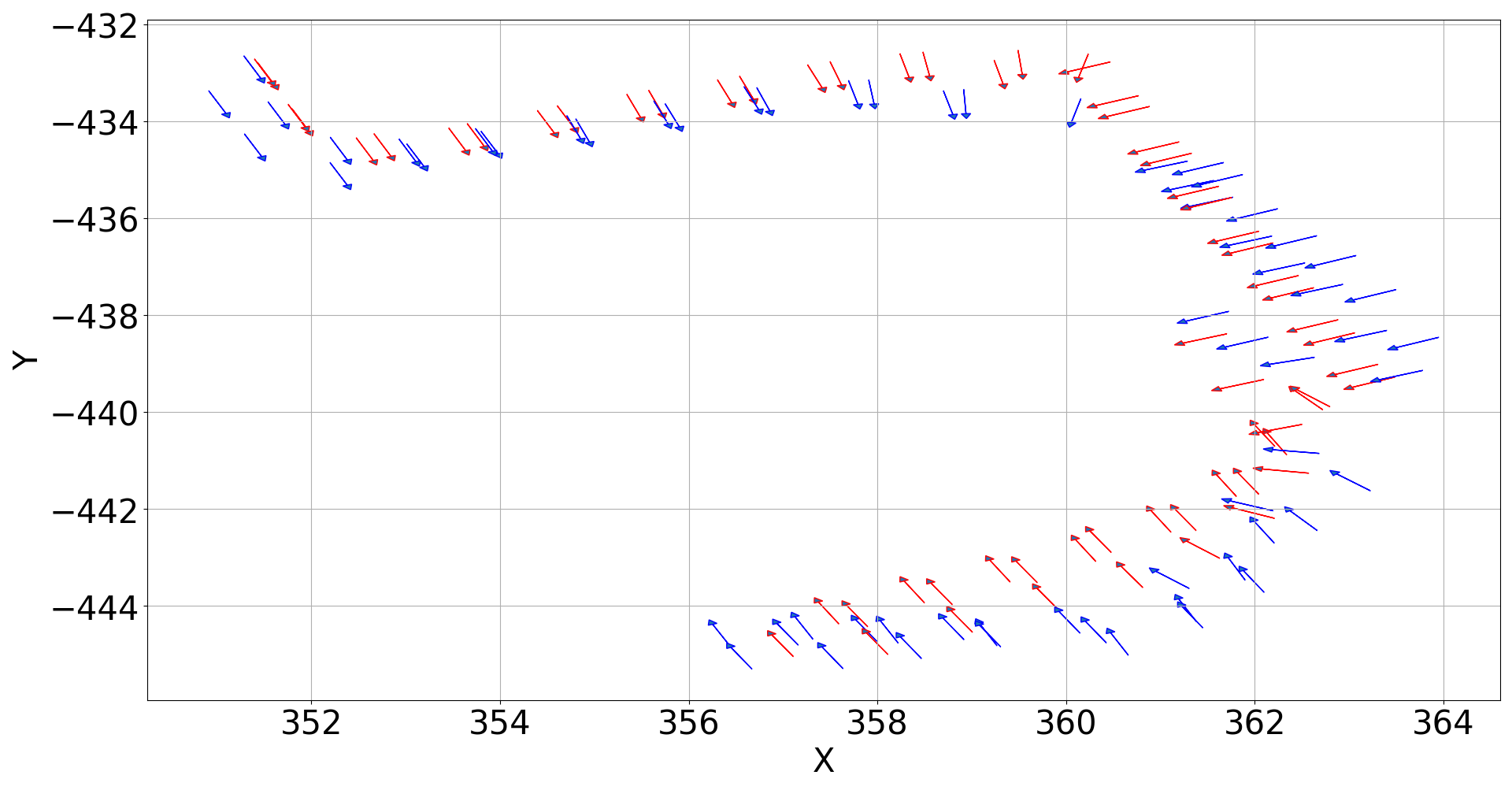}
    \end{minipage}
    \begin{minipage}[t]{0.40\textwidth}
        \centering
        \includegraphics[width=\textwidth]{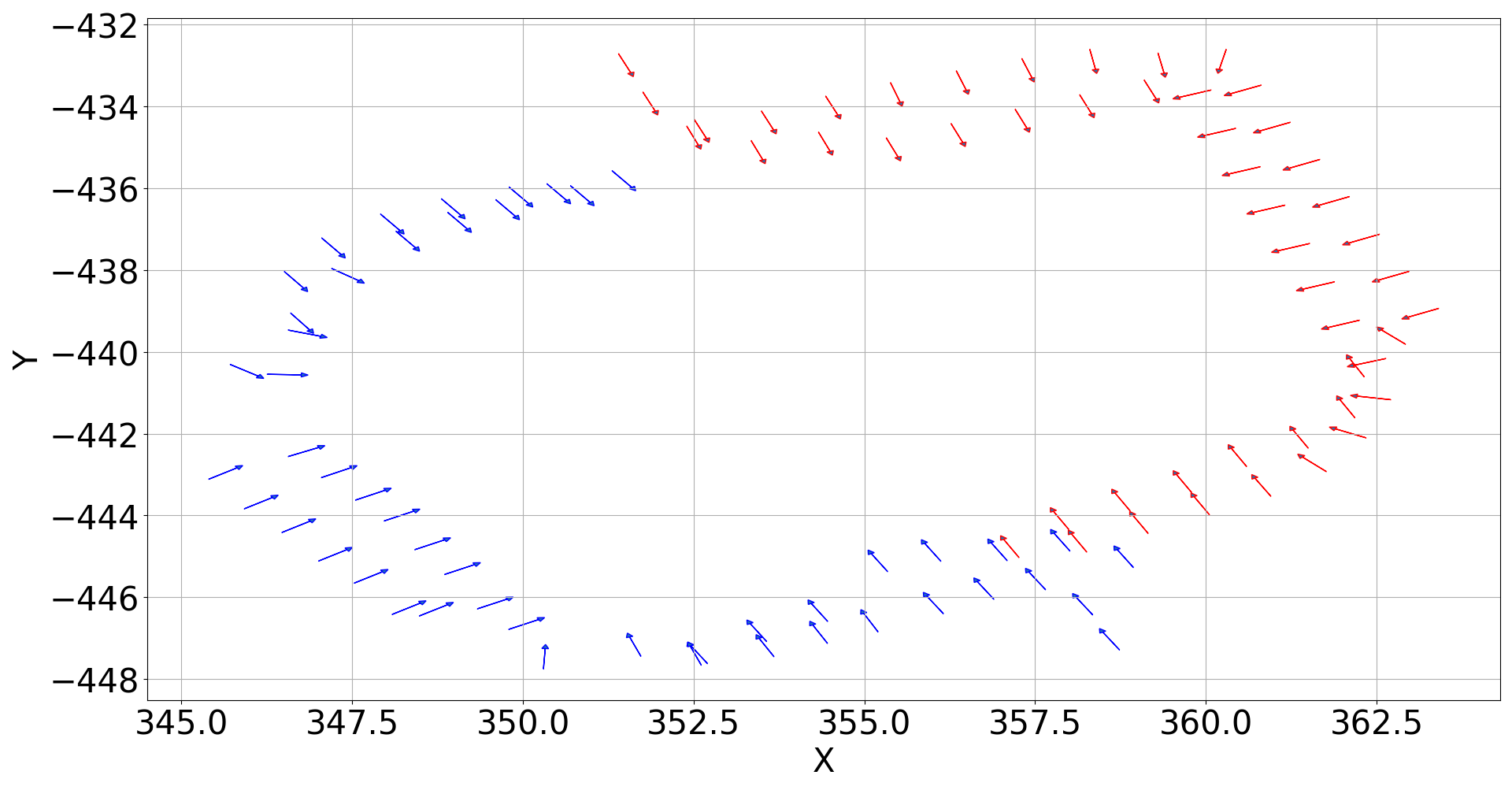}
    \end{minipage}
    \label{fig:loc_missions}
    \vspace{-5mm}
\end{figure*}

\section{Experiments}
\label{sec:experiments}
The experiments were carried out in a virtual environment, that was built using both Gazebo and the underwater plugin Dave. 
The latter was used to simulate the underwater currents, the AUV with its actuators, the INS sensors, and the Front-Looking sonar. Testing in virtual settings allows us to also evaluate quantitatively the accuracy and robustness of the proposed method because the ground truth information is available, plus the fact that we are able to control the noise that is injected into the simulation.
We evaluate the proposed method on two different modalities: 
\begin{itemize}
    \item Localization: We measure how fast and how precisely the robot from a lost state is able to localize itself in the underwater plant;
    \item Tracking: We track the evolution of the uncertainty of the robot's pose estimate given that the initial pose of the robot is known;
\end{itemize}
The pipeline has been evaluated only using the SAD module, which will be referred to as PF-SAD, and using the combination of the SAD and PRec modules, which will be referred to as PF-(SAD+PRec).
The methods have been tested on three different types of missions, that summarize all the possible different scenarios that can arise during a real inspection mission. The trajectories registered are all based on real missions.
The core of the Particle Filter was implemented in C++, while the observation modules were implemented in Python/Pytorch. All the modules and the simulations were run on a laptop with CPU 11th Gen Intel(R) Core(TM) i7-11800H @ 2.30G and GPU NVIDIA GeForce RTX 3060.

\begin{figure}[b!]
    \centering
    \begin{minipage}[t]{.23\textwidth}
        \centering
        \includegraphics[width=\textwidth]{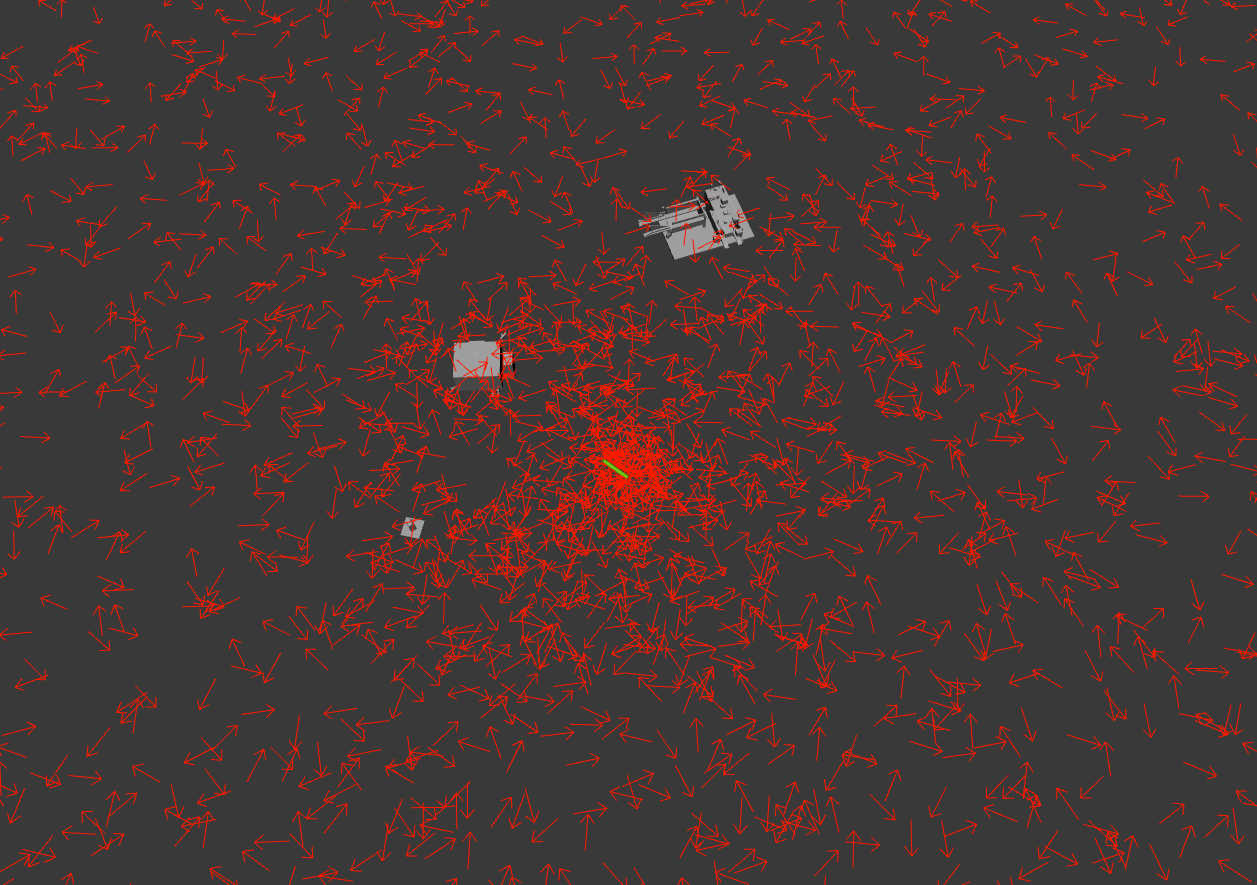}
    \end{minipage}
    \begin{minipage}[t]{.23\textwidth}
        \centering
        \includegraphics[width=\textwidth]{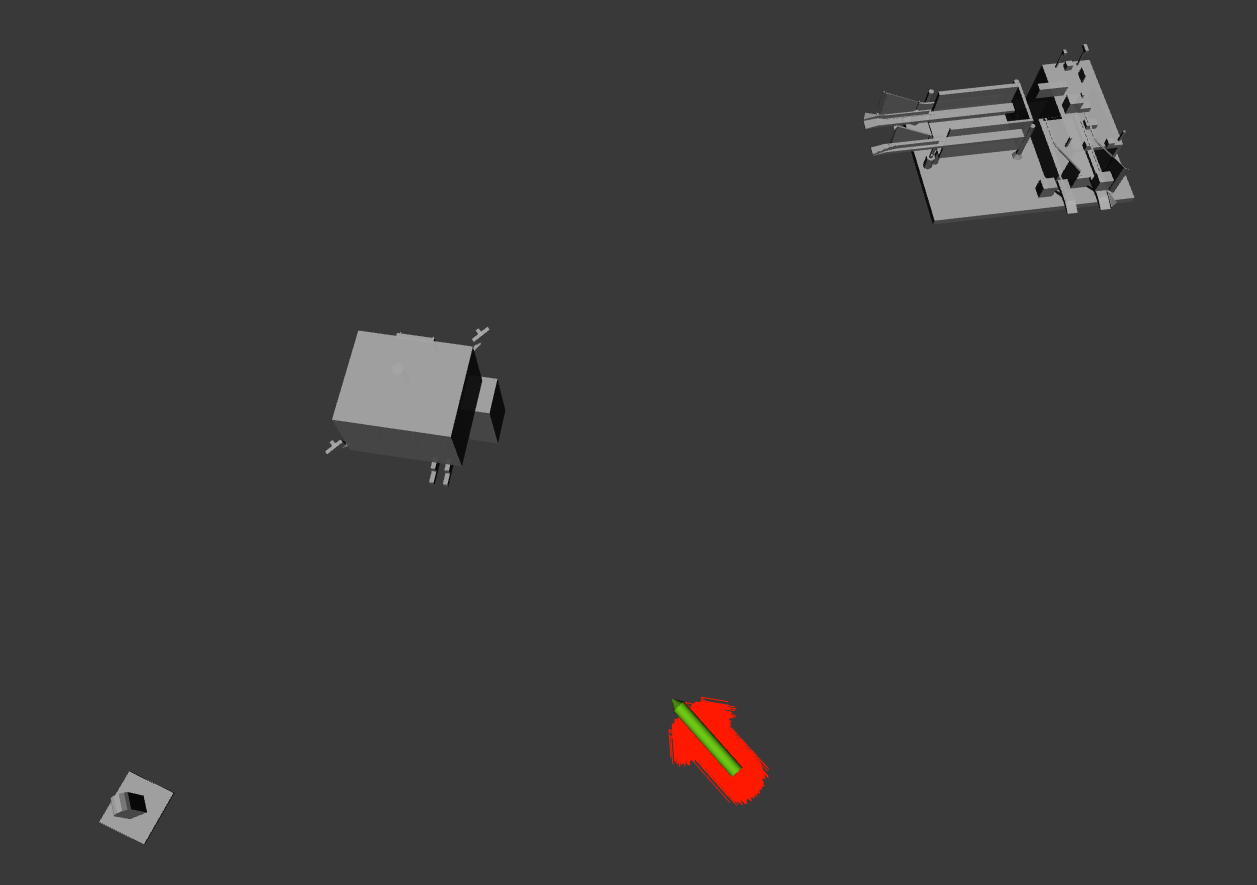}
    \end{minipage}
    \caption{The red arrows represent the particles, while the green arrows represent the ground-truth position of the AUV. The left and right images show, respectively, the distribution of the particles when localizing and when tracking the pose of the robot.}
    \label{fig:rviz}
\end{figure}

\begin{figure*}[t!]
    \centering
    \caption{The plots represent the evolution of the error during tracking. On the left column, it represented the position error(expressed in meters), while on the right column the angle error(expressed in radians). From top to bottom, they are mission 1, 2, 3 plots. The orange shade represents the variance in the error computed on different runs. For the angle error plots, this area is almost not noticeable.}
    \begin{minipage}[t]{0.4\textwidth}
        \centering
        \includegraphics[width=\textwidth]{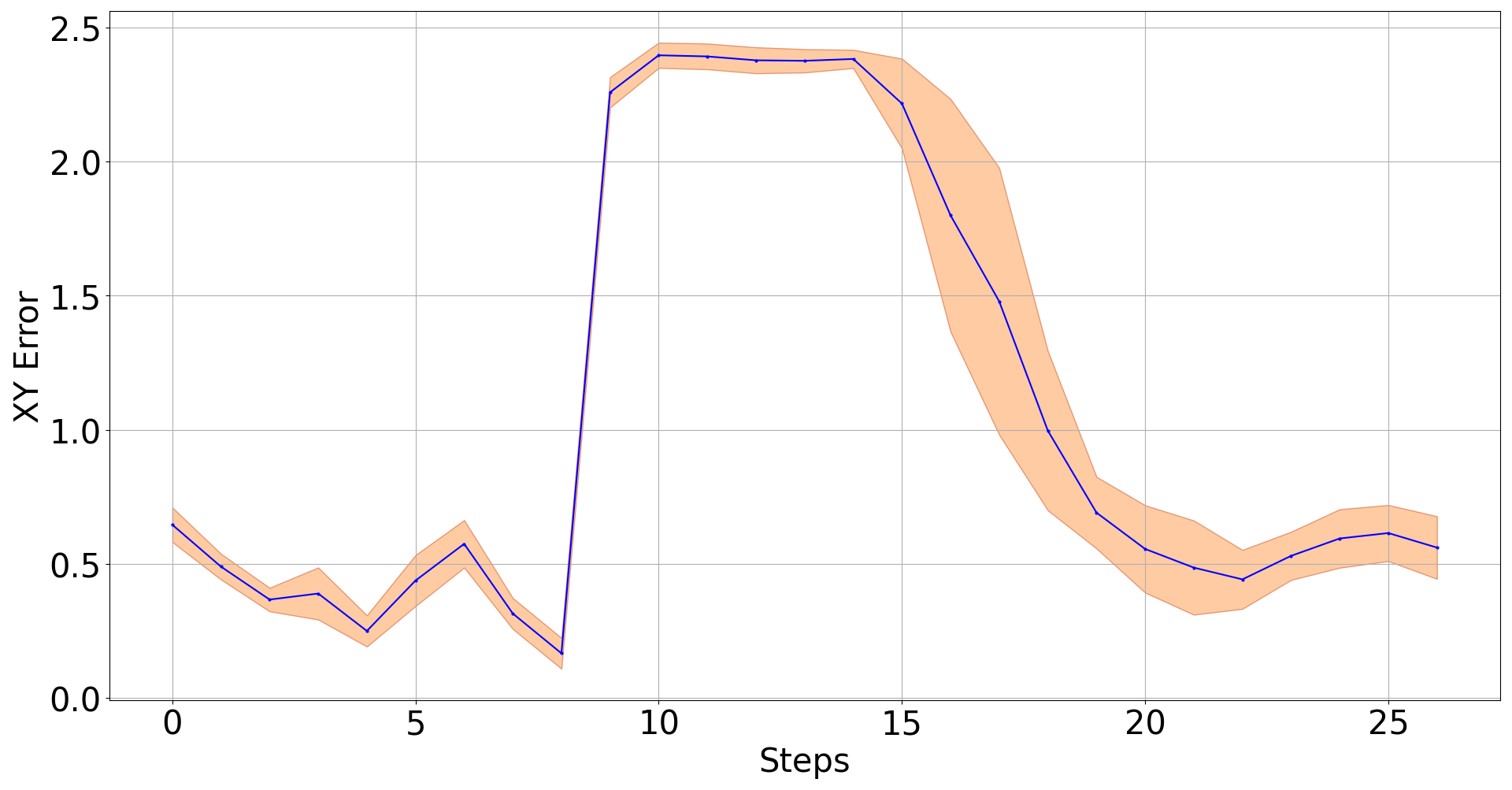}
    \end{minipage}
    \begin{minipage}[t]{0.4\textwidth}
        \centering
        \includegraphics[width=\textwidth]{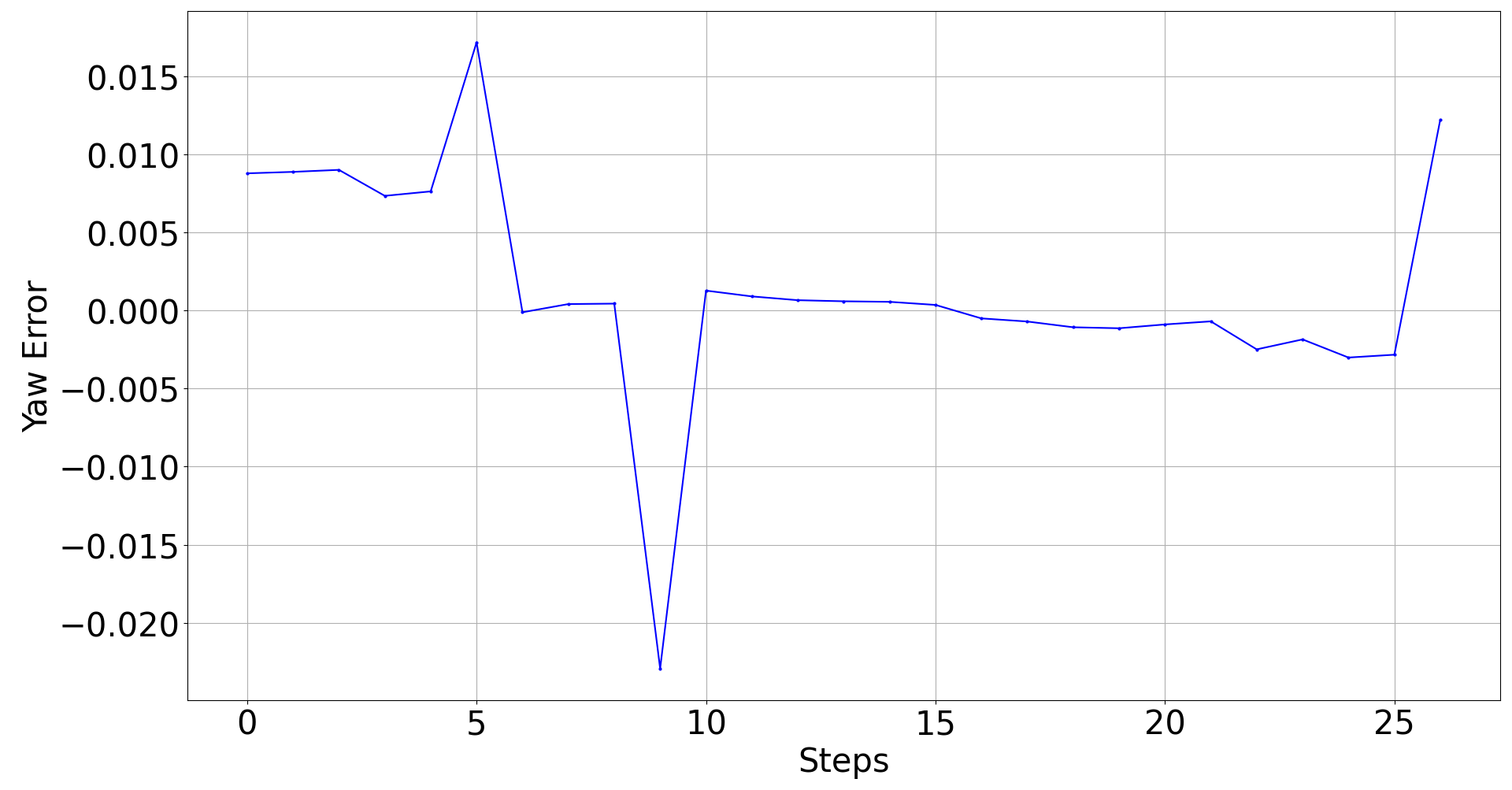}
    \end{minipage}
    \begin{minipage}[t]{0.4\textwidth}
        \centering
        \includegraphics[width=\textwidth]{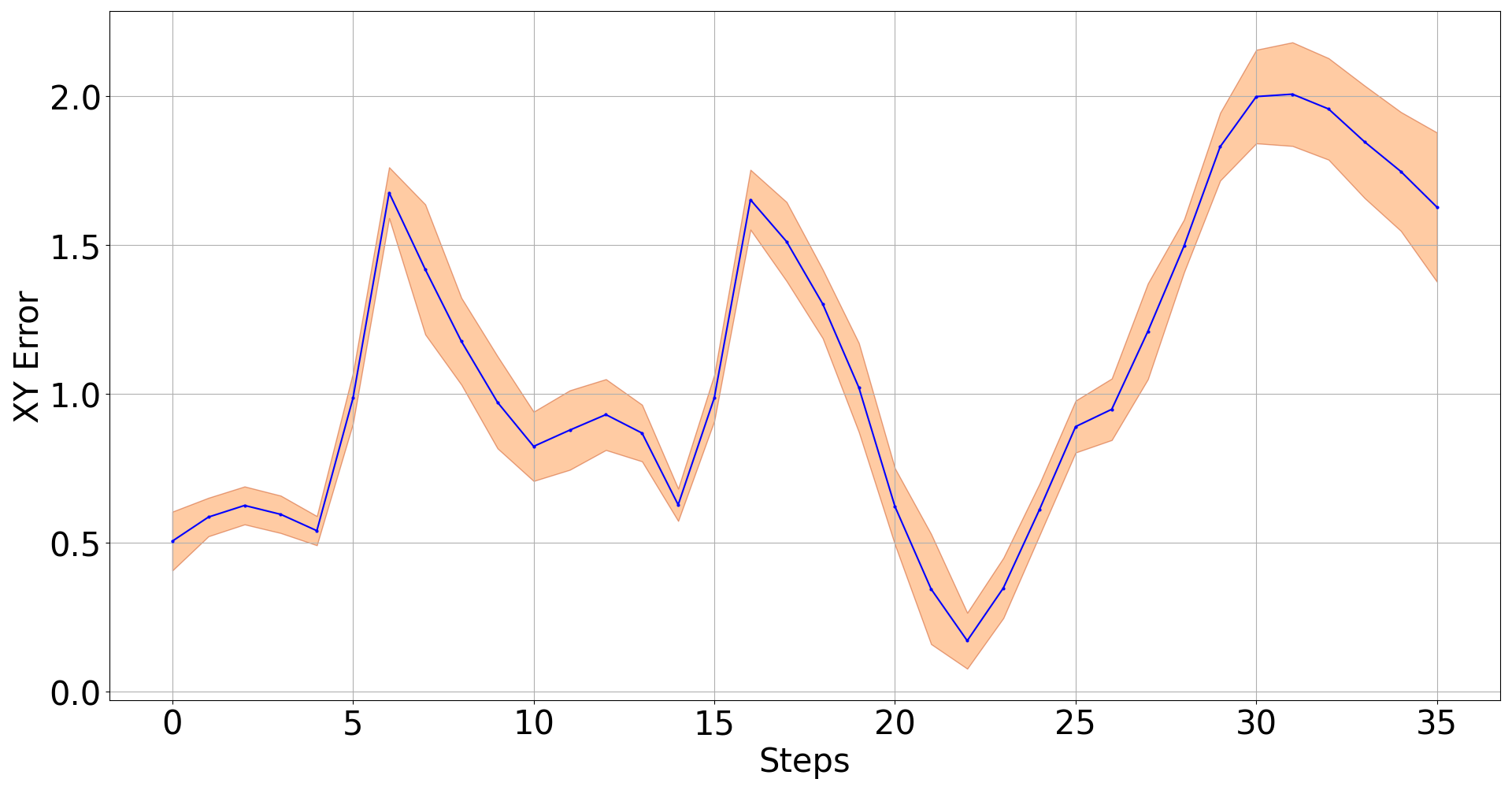}
    \end{minipage}
    \begin{minipage}[t]{0.4\textwidth}
        \centering
        \includegraphics[width=\textwidth]{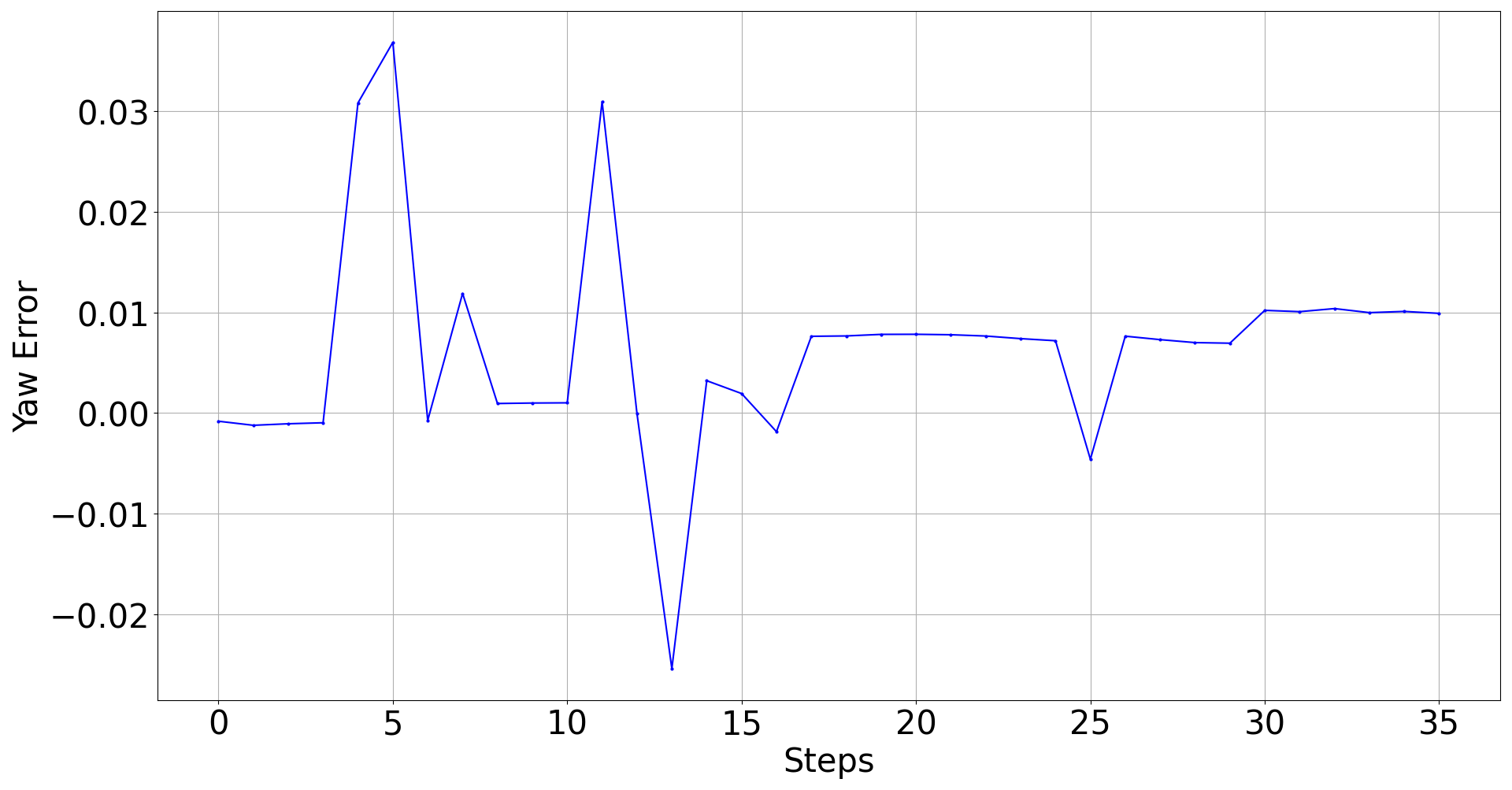}
    \end{minipage}
    \begin{minipage}[t]{0.4\textwidth}
        \centering
        \includegraphics[width=\textwidth]{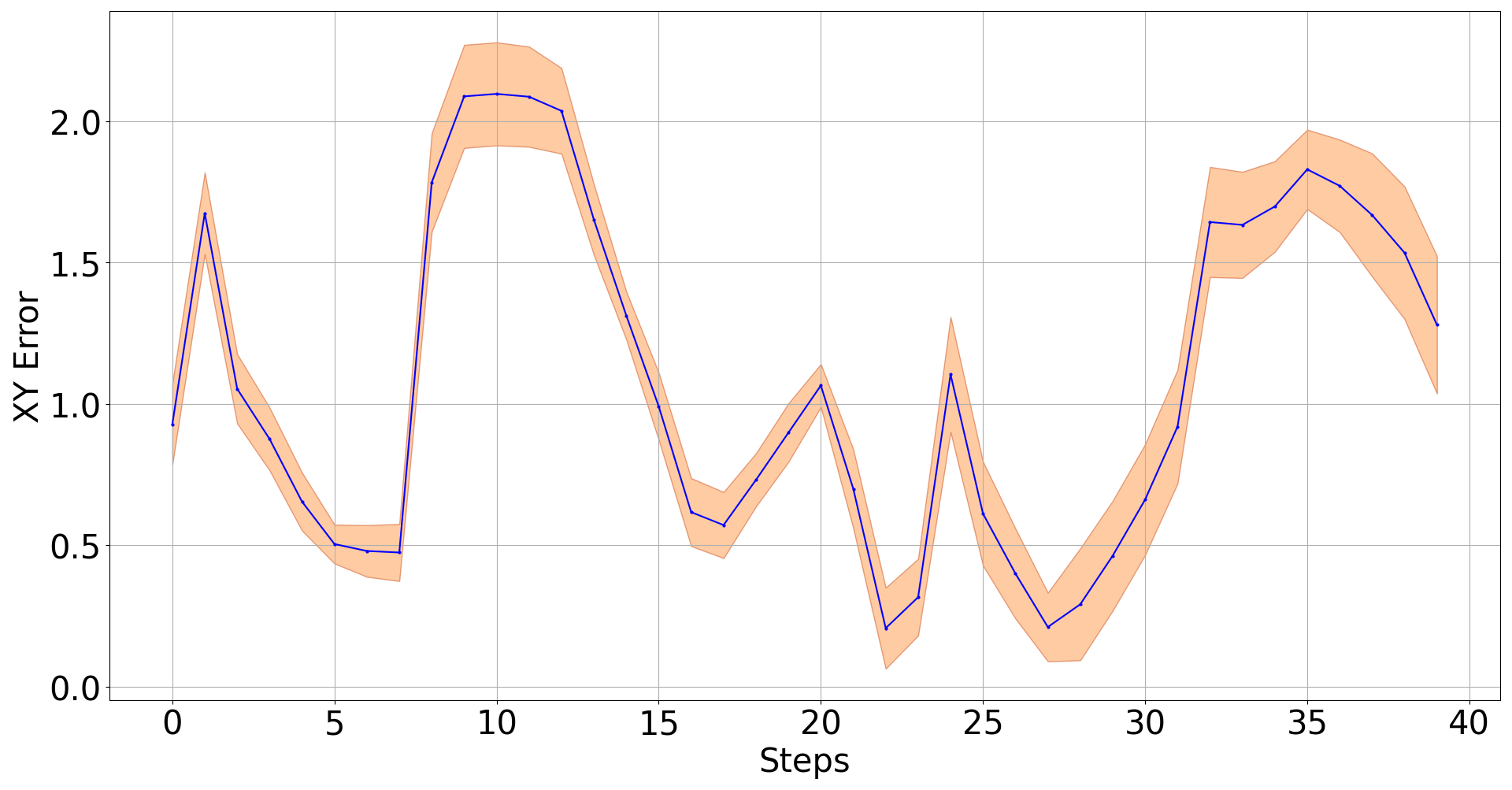}
    \end{minipage}
    \begin{minipage}[t]{0.4\textwidth}
        \centering
        \includegraphics[width=\textwidth]{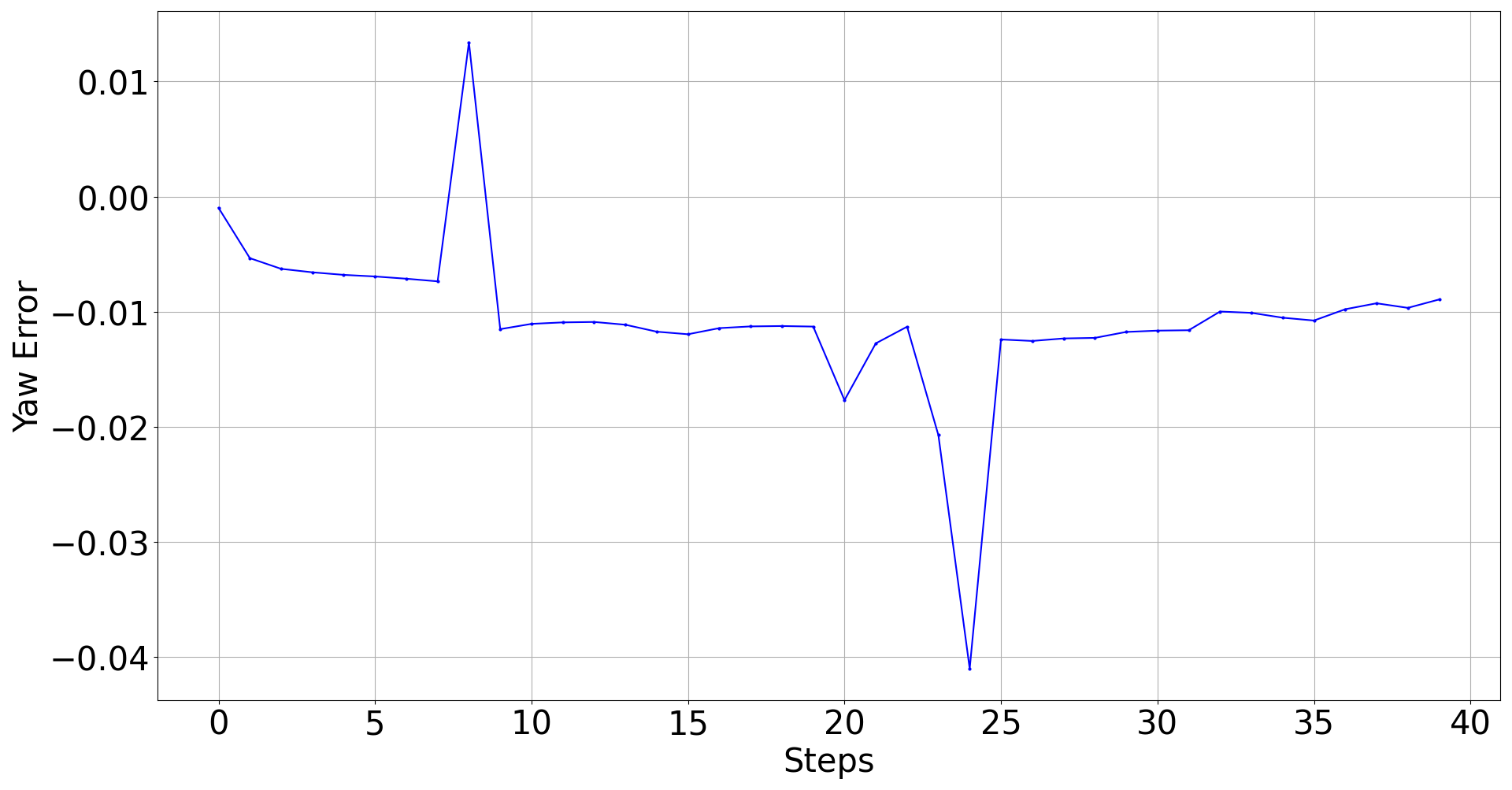}
    \end{minipage}
    \label{fig:tracking_missions}
\end{figure*}

\subsection{Testing Environment}
The scene developed for testing is populated with 4 underwater assets: two of them are within $\sim$20m distance between one another, while others are $\sim$100m apart. 
Dave not only simulates the sonar sensors but also allows to personalize their characteristics. We used this functionality to set the sonar parameters to be the same as the ones of the real FLS used in their inspection missions, which is the Tritech Gemini 720i Multibeam Imager. It was set to have $R_{max} = 20m$, $FoV = 120^\circ$ and to produce a sonar image with width $W = 440$ and height $H = 512$. 

\begin{table*}[t]{}
\centering
\resizebox{0.7\linewidth}{!}{
\begin{tabular}{|c|ccl|ccl|ccl|}
\hline
\multirow{2}{*}{} &
  \multicolumn{3}{c|}{Mission 1} &
  \multicolumn{3}{c|}{Mission 2} &
  \multicolumn{3}{c|}{Mission3} \\ \cline{2-10} 
 &
  \multicolumn{1}{c|}{\begin{tabular}[c]{@{}c@{}}Success \\ rate(\%)\end{tabular}} &
  \multicolumn{2}{c|}{Acc(m)} &
  \multicolumn{1}{c|}{\begin{tabular}[c]{@{}c@{}}Success \\ rate(\%)\end{tabular}} &
  \multicolumn{2}{c|}{Acc(m)} &
  \multicolumn{1}{c|}{\begin{tabular}[c]{@{}c@{}}Success \\ rate(\%)\end{tabular}} &
  \multicolumn{2}{c|}{Acc(m)} \\ \hline
PF-SAD &
  \multicolumn{1}{c|}{90} &
  \multicolumn{2}{c|}{0.49+ 0.69} &
  \multicolumn{1}{c|}{80} &
  \multicolumn{2}{c|}{0.73+ 0.75} &
  \multicolumn{1}{c|}{50} &
  \multicolumn{2}{c|}{4.95 + 4.84} \\ \hline
PF-SAD+PRec &
  \multicolumn{1}{c|}{100} &
  \multicolumn{2}{c|}{0.18 + 0.02} &
  \multicolumn{1}{c|}{100} &
  \multicolumn{2}{c|}{0.24 + 0.004} &
  \multicolumn{1}{c|}{100} &
  \multicolumn{2}{c|}{0.34 + 0.023} \\ \hline
\end{tabular}
}
\caption{Comparison in the localization performances of the partial and full pipeline during the different missions.}
\label{tab:success_loc}
\vspace{-7mm}
\end{table*}

\subsection{Localization}
The localization experiments consist of uniformly distributing the particles in a circle of radius $r = 100$ meters in the X-Y plane centered in the initial ground-truth pose of the AUV. After identifying the position of the particle, also its orientation yaw is randomly set between $[0, 2\pi]$. The roll and pitch parameters are set to zero, while the $z$ coordinate of the vehicle is set to be the same as the one of the AUV for the assumptions stated in \secref{sec:assumptions}.
For the different missions, $N = 3000$ particles were generated in the environment to evaluate the speed of convergence of the filter and the accuracy of the position estimate. These factors were evaluated each time the vehicle navigates 1 meter from its previous position.
\subsection{Tracking}
The tracking experiments consist of uniformly distributing the particles in a circle of radius $r = 1$ meter in the X-Y plane centered in the initial ground-truth pose of the AUV. while in this case its yaw is set to be the same as the ground-truth one with a small random perturbation between $[-\frac{\pi}{20}, \frac{\pi}{20}]$. The same considerations as the ones used for Localization are used here for the rest of the parameters.
For the different missions, $N = 1500$ particles were generated to evaluate the accuracy of the position estimate and the amount of drift that has been eliminated. These factors were evaluated each time the vehicle navigates 1 meter from its previous position.
\subsection{Inspection Missions}
The different missions were numbered based on the difficulty level in terms of localization, from easiest (1) to hardest (3). The first mission represents the scenario in which the AUV has to inspect multiple close-by assets in a singular mission. In particular, the mission, shown in the top left of \figref{fig:loc_missions}, consists of the inspection of two nearby assets. Even in the case that ambiguity could arise from the SAD module during the inspection of the first module, it then is eliminated when inspecting the second one. In fact, as we can see from the first column of \tabref{tab:success_loc} have almost the same performances in both success rate and accuracy in terms of localization.\\
The second mission represents the scenario in which the AUV has to inspect a singular far-away asset, that is not strongly symmetric in the sonar image plane. It can be seen, from the second column of \tabref{tab:success_loc}, that even in cases of not strong symmetry the amount of successful runs starts to diminish when using only the SAD, while also for mission 2 the SAD+PRec the results are stable both in terms of success rate and accuracy.
The final and third mission is the hardest one in terms of localization because it consists of inspecting a singular symmetric asset. In fact, the third column of \tabref{tab:success_loc} shows exactly what would be expect given the observation model used in SAD, 50\% of success rate due to the ambiguity in the measurement. In the bottom right of \figref{fig:loc_missions}, it is displayed an instance of a failed localization during mission 3. Instead, using SAD+PRec in the localization allows having always successful localization with stable accuracies even in the most challenging scenarios. Instead, the tracking results between the two different modalities SAD and SAD+PRec were equivalent, as it makes sense, being that the PRec module adds information to SAD only when there are ambiguous situations. In fact, only the performances of SAD+PRec were reported in \figref{fig:tracking_missions} to avoid redundancy. From top to bottom, it is displayed the tracking error evolution during missions 1, 2, and 3, while from left to right it is displayed the position error in the X-Y plane, and the angle error. The angle estimation flutters maximum between $[-0.041, 0.034]$ radians, while the position error in the X-Y plane varies in the range $[0.18, 2.87]$ meters in all the different missions. Proving that the proposed method is able to handle all the different types of scenarios that could arise when carrying out underwater inspection operations, at least, in virtual settings.
The different observation models were tested using real sonar images coming from the modeled underwater plant, but the gap from simulation to real measurement was too wide and the observation models were only able to provide spurious detections. 

\section{Conclusions}
\label{sec:conclusions}
We propose a novel underwater real-time localization algorithm based on Particle Filter. The solution has been tailored to work in sparsely populated underwater environments for inspection missions. The proposed solution doesn't require a rich map, but it only needs as input the poses of the underwater assets with respect to a pre-defined reference frame.
The method was preliminarily evaluated in a simulated environment, with promising results.\\
Future work will focus on developing a real-to-sim network to bridge the gap between real and simulated sonar measurements. Transitioning from real to simulation, rather than performing the inverse step, will yield cleaner and simpler measurements, enhancing the performances of the observations model on real data, while also avoiding the need for multiple expensive acquisition campaigns.

\balance
{
\bibliographystyle{plain}
\bibliography{references}
}

\end{document}